\documentclass{article}

\usepackage[utf8]{inputenc} % allow utf-8 input
\usepackage[T1]{fontenc}    % use 8-bit T1 fonts
\usepackage{hyperref}       % hyperlinks
\usepackage{url}            % simple URL typesetting
\usepackage{booktabs}       % professional-quality tables
\usepackage{amsfonts}       % blackboard math symbols
\usepackage{nicefrac}       % compact symbols for 1/2, etc.
\usepackage{microtype}      % microtypography
\usepackage{xcolor}         % colors

\usepackage[nonatbib, preprint]{neurips_2025}
\usepackage{biblatex}
\usepackage{amsmath}
\usepackage{amsthm}
\usepackage{dsfont}
\usepackage{microtype}
\usepackage{adjustbox}
\usepackage{graphicx}
\usepackage{subcaption}
\usepackage{float}
\usepackage{multirow}
\usepackage{makecell}
\usepackage{booktabs} % for professional tables
\usepackage[flushleft]{threeparttable}
\usepackage{pythonhighlight}

\newtheorem{proposition}{Proposition}
\newtheorem{definition}{Definition}
\newtheorem{theorem}{Theorem}
\newtheorem*{Proof}{Proof}
\bibliography{references}
\usepackage{comment}
\allowdisplaybreaks %allows page breaks %in \alogn
\title{ROCK: Riesz Occupation Kernel Methods for Learning Dynamical Systems}

\author{%
  Victor W Rielly\\
  Department of Mathematics\\
  Portland State University\\
  Portland, OR 97206 \\
  \texttt{victor23@pdx.edu} \\
  \And
   Chau Nguyen \\
   Portland State University \\
   \texttt{min25@pdx.edu} \\
   \AND
   Kamel Lahouel \\
   Translational Genomics Research Institution\\
   \texttt{klahouel@tgen.org} \\
   Anthony Kolshorn \\
   Portland State University \\
   \texttt{kolshorn@pdx.edu} \\
   \And
   Nicholas Fisher \\
   Portland State University \\
  \texttt{nicholfi@pdx.edu} \\
  \And
  Bruno Jedynak \\
  Portland State University\\
  \texttt{bjedyna2@pdx.edu}\\
}

\begin{document}

\maketitle

\begin{abstract}
We present a representer theorem for a large class of variational problems, generalizing regression, finite element methods, and providing techniques for learning partial and ordinary differential equations (PDEs and ODEs). We apply our formulation to the multivariate occupation kernel method (MOCK) for learning dynamical systems from data. Our generalized method outperforms the MOCK method on most tested datasets, while often requiring considerably fewer parameters. Experiments are presented for learning ODEs and PDEs. 
\end{abstract}

%There is a deep connection between dynamical systems and deep learning models, as was highlighted by \cite{neuralode} and follow-up papers such as \cite{pinode}. Indeed, a forward pass through a deep model may be thought of as a discrete realization of a trajectory in a dynamical system. It stands to reason that efficient learning paradigms for dynamical systems could be used to make more efficient learning paradigms for deep neural networks and vice versa. We present a general learning paradigm for variational problems and apply this paradigm to develop a state-of-the-art method for learning dynamical systems. 
\section{Introduction}

%\subsection{Problem formulation}

Many statistical learning problems can be cast in the form \cite{vapnik1991principles}
\begin{equation}
\min_{f \in \mathcal{F}} \left\{ \sum_{i=1}^N L(f(x_i), y_i) + \Omega(f) \right\},
\end{equation}
where \( \mathcal{F} \) is a suitable hypothesis space, \( y_i \) are observed responses, and \( \Omega(f) \) is a regularization term. The evaluation of \( f \) at \( x_i \) is a linear functional of \(f\). This naturally leads to a generalized formulation:
\begin{equation}
\min_{f \in \mathcal{F}} \left\{ \sum_{i=1}^N L(a_i(f), y_i) + \Omega(f) \right\},
\end{equation}
where \( a_i(f) \) denotes a linear functional applied to \( f \), possibly varying with \( i \).

A further generalization incorporates adversarial, test function-based estimation, in which we replace the linear functionals $a_i$ with bilinear functionals involving $f$ and test functions $v_i$, leading to the variational problem:
\begin{equation}
\min_{f \in \mathcal{F}} \left\{ \sum_{i=1}^N \sup_{v_i \in \mathds{V}_i| \|v_i\|=1} \left\{ L(a_{i}(v_i,f), l_i(v_i)) \right\} + \Omega(f) \right\}.
\end{equation}
Each \( \mathds{V}_i \) is a space of test functions, \( l_i(\cdot) \) is a linear functional on \( \mathds{V}_i \), and \( a_{i}(v_i,f) \) is a bilinear functional. Notice that to make the supremum well defined, we need to optimize over test functions of unit norm.

In this work, we focus on the quadratic case where \( L(x, y) = (x - y)^2 \), and consider regularization of the form \( \Omega(f) = \lambda \|f\|^2 \). Remarkably, under the assumption that \( \mathcal{F} \) is a vector-valued reproducing kernel Hilbert space (vvRKHS), this problem admits a closed-form solution. 

We demonstrate that this framework captures a broad range of learning problems, including multivariate ridge regression; learning the vector fields of ordinary differential equations (ODEs) from data; and learning certain classes of partial differential equations (PDEs). For instance, we can learn PDEs of the form
\[
\frac{\partial u}{\partial t} = f\left( \partial_x u, \partial_{xx} u, \ldots \right),
\]
where the time derivative depends explicitly on spatial derivatives of \( u \). Such equations arise in heat diffusion, reaction-diffusion systems, and other spatiotemporal processes. These examples highlight the versatility and expressiveness of our variational framework.

\subsection{Related work}

VvRKHSs have received increasing attention in recent years. A particularly relevant work is \cite{minh2016unifying}, which also presents a unified framework for learning in vvRKHSs. There is a rich body of literature on kernel-based methods for learning and solving PDEs and dynamical systems, especially through the use of collocation points \cite{ChenFuChen}. Our approach is closely aligned with the perspective of optimal recovery. For instance, in \cite{owhadi2019operator}, the central problem formulated in Chapter 3 for optimal recovery splines involves a minmax problem similar to ours. %In our case, the supremum is taken over functions of unit norm, whereas in \cite{owhadi2019operator}, the objective involves minimizing the ratio of a loss to the norm of the function, which is an equivalent formulation in the case of linear
Additional work in this direction includes \cite{long2023kernelapproachpdediscovery}, which exhibits several parallels with our framework. By formulating our results within the vvRKHS setting, we are able to exploit the computational advantages of separable kernels, as detailed in \cite{rielly2023learning}.

\subsection{Vector-valued reproducing kernel Hilbert spaces }

The central tool that enables analytical tractability in our framework is the structure of vvRKHSs. We assume both the hypothesis function \( f \) and the test functions \( v_i \) belong to suitable vvRKHSs. This allows us to exploit the linearity of evaluation functionals, derive closed-form solutions, and obtain efficient representations of our estimator of $f$. We briefly recall the definition of vvRKHSs below. For a thorough exposition on reproducing kernel Hilbert spaces, see \cite{paulsen2016introduction}.
\begin{definition}
Let $\mathds{H}$ be a Hilbert space of vector-valued functions taking a nonempty set $\mathcal{X}$ to $\mathds{R}^d$. $\mathds{H}$ is a vvRKHS if for every $v \in \mathds{R}^d$ and $x \in \mathcal{X}$, the linear functional
\begin{equation}
l_{v,x}(f) \equiv f(x)^T v
\end{equation}
is bounded.
\end{definition} 
A more complete definition is provided in Appendix \ref{app: vvrkhs}. As a trivial yet interesting example, every finite-dimensional Hilbert space of vector-valued functions is a vector-valued RKHS.

\section{Riesz occupation kernel method}
%\subsection{Problem statement and solution}
The representer theorem is a foundational result in machine learning \cite{Schlokopf}. It states that some functional optimization problems have a unique finite-dimensional solution as a linear combination of data-dependent functions. In the quadratic case, this solution is obtained by solving a linear system. We provide a generalization of this result, extending its applicability. The proof uses standard functional analysis tools, including the Riesz representation theorem and orthogonal projections. This proof is relegated to Appendix \ref{app:thm1_2}.       
\begin{theorem}[Representer theorem]
\label{thm:BigThm_sec}
Let $\mathds{H},\mathds{V}_1,...,\mathds{V}_n$ be vvRKHSs with kernels $K_{\mathds{H}},K_{\mathds{V}_1},...,K_{\mathds{V}_n}$, inner products $\left<\cdot,*\right>_{\mathds{H}},\left<\cdot,*\right>_{\mathds{V}_1},...,\left<\cdot,*\right>_{\mathds{V}_n}$ and norms $\|\cdot\|_{\mathds{H}},\|\cdot\|_{\mathds{V}_1},...,\|\cdot\|_{\mathds{V}_n}$ respectively. Assume $\mathds{H}$ takes a set $\mathcal{X}$ to $\mathds{R}^d$ and $\mathds{V}_i$ takes a set $\mathcal{Y}_i$ to $\mathds{R}^{d_i}$. Assume also $\mathds{V}_1,...,\mathds{V}_n$ are finite dimensional with dimensions $q_1,...,q_n$ and feature functions $\Psi_1,...,\Psi_n$ with $\Psi_i: \mathcal{Y}_i \Rightarrow \mathds{R}^{q_i\times d_i}$  respectively. Let $p_1,...,p_n$ be bilinear functionals taking $\mathds{H} \times \mathds{V}_i$ to $\mathds{R}$ that are continuous with respect to both entries. Let $l_1,...,l_n$ be linear functionals taking $\mathds{V}_1,...,\mathds{V}_n$ to $\mathds{R}$ that are also continuous. We notate $e_j$ the $j^{th}$ element of the natural basis of the appropriate Euclidean space.  Then, the variational problem \begin{equation}
\label{eq:Loss}
\inf_{f\in\mathds{H}}\left\{\sum_{i=1}^{n}\sup_{v\in\mathds{V}_i,\|v\|=1}\left\{\left(p_{i}(f,v)-l_i(v)\right)^2\right\} + \lambda\|f\|^2_{\mathds{H}}\right\}
\end{equation}
admits a unique solution. Moreover, this solution is provided by solving the linear system
\begin{comment}
That is to say, for a fixed $v\in\mathds{V}_i$, There is a constant $K_v$ such that for any $h\in\mathds{H}$. %\comment[vr]{Reword}
\begin{equation}
|p_i(h,v)| \leq K_v\|h\|_{\mathds{H}}
\end{equation}
and for any fixed $h\in \mathds{H}$ there is a constant $K_h$ such that for any $v\in \mathds{V}_i$
\begin{equation}
|p_i(h,v)|\leq K_h\|v\|_{\mathds{V}_i}
\end{equation}
Let $l_1,...,l_n$ be continuous linear functionals taking $\mathds{V}_1,...,\mathds{V}_n$ to $\mathds{R}$ so that there is a constant $K_i$ such that for any $v\in\mathds{V}_i$, 
\begin{equation}
|l_i(v)|\leq K_i\|v\|_{\mathds{V}_i}
\end{equation}
(Note, since $\mathds{V}_i$ are finite dimensional, all linear functionals taking $\mathds{V}_i$ to $\mathds{R}$ are bounded)
then for any strictly increasing function $\lambda : [0,\infty) \rightarrow [0,\infty) $ there is a unique solution to the variational problem:
\begin{equation}
\label{eq:Loss}
\inf_{f\in\mathds{H}}\left\{\sum_{i=1}^{n}\sup_{\left\{v\in\mathds{V}_i|\|v\|=1\right\}}\left\{\left(p_{i}(f,v)-l_i(v)\right)^2\right\} + \lambda(\|f\|^2_{\mathds{H}})\right\}
\end{equation}
Moreover, assuming $\lambda(\|x\|_{\mathds{H}}^2) = \lambda\cdot \|x\|_{\mathds{H}}^2, \lambda > 0$ (is the quadratic function) this unique solution is given by solving the linear system
\end{comment}
\begin{equation}
\label{eq:linear_system}
\left(M+\lambda I\right) \alpha = y,
\end{equation}
where $\forall x \in \mathcal{X}$, $f(x)$ is  given by
\begin{equation}
f(x)= \sum_{i=1}^{n}\sum_{k=1}^{q_i}\alpha_{i,k}\mathcal{L}^{*}_{i,k}(x), \  \mathcal{L}^*_{i,k}(x) = \sum_{j=1}^{d}p_i(K_{\mathds{H}}(\cdot,x)e_j,\Psi_i(\cdot)^Te_k)e_j.
\end{equation}
\begin{comment}
where
\begin{equation}
\mathcal{L}^*_{i,k}(*) = \sum_{j=1}^{d}p_i(K_{\mathds{H}}(\cdot,*)e_j,\Psi_i(\cdot)^Te_k)e_j
\end{equation}
\end{comment}
Letting $N = \sum_{i=1}^{n}q_i$, $M\in \mathds{R}^{N\times N}$ is given by
%\begin{equation}
$M_{l,m:i,k} = \left<\mathcal{L}^*_{l,m},\mathcal{L}^*_{i,k}\right>_{\mathds{H}}$
%\end{equation}
so that
%\begin{small}
%\begin{equation*}
%    M = \left[\begin{array}{cccccccccc}

%M_{1,1,1,1}&...&M_{1,1,1,q_1}&M_{1,1,2,1}&...&M_{1,1,2,q_2}&...&M_{1,1,n,1}&...&M_{1,1,n,q_n} \\
%    \vdots& \ddots& \vdots& \vdots& \ddots& \vdots& \ddots& \vdots& \ddots& \vdots \\
%M_{1,q_1,1,1}&...&M_{1,q_1,1,q_1}&M_{1,q_1,2,1}&...&M_{1,q_1,2,q_2}&...&M_{1,q_1,n,1}&...&M_{1,q_1,n,q_n}\\
%M_{2,1,1,1}&...&M_{2,1,1,q_1}&M_{2,1,2,1}&...&M_{2,1,2,q_2}&...&M_{2,1,n,1}&...&M_{2,1,n,q_n}\\
%\vdots& \ddots& \vdots& \vdots& \ddots& \vdots& \ddots& \vdots& \ddots& \vdots \\
%M_{2,q_2,1,1}&...&M_{2,q_2,1,q_1}&M_{2,q_2,2,1}&...&M_{2,q_2,2,q_2}&...&M_{2,q_2,n,1}&...&M_{2,q_2,n,q_n}\\
%\vdots& \ddots& \vdots& \vdots& \ddots& \vdots& \ddots& \vdots& \ddots& \vdots \\
%M_{n,1,1,1}&...&M_{n,1,1,q_1}&M_{n,1,2,1}&...&M_{n,1,2,q_2}&...&M_{n,1,n,1}&...&M_{n,1,n,q_n}\\
%\vdots& \ddots& \vdots& \vdots& \ddots& \vdots& \ddots& \vdots& \ddots& \vdots \\
%M_{n,q_n,1,1}&...&M_{n,q_n,1,q_1}&M_{n,q_n,2,1}&...&M_{n,q_n,2,q_2}&...&M_{n,q_n,n,1}&...&M_{n,q_n,n,q_n}\\
%    \end{array}\right]
%\end{equation*}
%\end{small}
\begin{equation*}
M = \left[\begin{array}{ccc}
M_{1:1}&...&M_{1:n} \\
\vdots&\ddots&\vdots \\
M_{n:1}&...&M_{n:n}
\end{array}\right]
%\end{equation*}
%where the $i,j$th block is the matrix:
\mbox{ with }
%\begin{equation*}
M_{i:j} = \left[\begin{array}{ccc}
    M_{i,1:j,1}&...&M_{i,1:j,q_j} \\
    \vdots &\ddots & \vdots \\
    M_{i,q_i:j,1}&...&M_{i,q_i:j,q_j}
\end{array}\right]
\end{equation*}
with
\begin{equation}
M_{l,m:i,k} = \left<\mathcal{L}^*_{l,m},\mathcal{L}^*_{i,k}\right>_{\mathds{H}}=p_i\left(\sum_{j=1}^{d}p_l\left(K_{\mathds{H}}(\cdot,*)e_j,\Psi_{l}(\cdot)^Te_m\right)e_j,\Psi_{i}(*)^Te_k\right)
\end{equation}
where $* \mapsto p_l\left(K_{\mathds{H}}(\cdot,*)e_j,\Psi_{l}(\cdot)^Te_m\right) e_j $ is viewed as a function in $\mathds{H}$, taking * as an input, 
and $y$ is given by
\begin{equation}
y_{i,k} = l_i(\Psi_i(\cdot)^Te_k)
%\end{equation}
\mbox{ so that } 
%\begin{equation}
y^T = \left[
y_{1,1},
...,
y_{1,q_1},
y_{2,1},
...,
y_{2,q_2},
...,
y_{n,q_n}
\right]
\end{equation}
\end{theorem}

It should be emphasized that all parametric methods are inherently finite-dimensional, and all linear functionals on finite-dimensional spaces are bounded. Examples of common linear functionals that arise in defining the loss in \eqref{eq:Loss} include function evaluation, integration, differentiation, and expected values.
\subsection{Example applications:}
To emphasize the breadth of applications, we present some problems that are interpretable within this framework.

\textbf{Multivariate ridge regression:} minimizing the loss
\begin{equation}
\min_{f}\left\{\sum_{i=1}^{n}\|f(x_i) - y_i\|_{\mathds{R}^d}^2 + \lambda\|f\|^2_{\mathds{H}}\right\}
\end{equation}
can be expressed by
\begin{equation}
\min_{f\in\mathds{H}}\left\{\sum_{i=1}^{n}\sum_{j=1}^{d}\sup_{v\in\{e_j,-e_j\}}\left\{(v^Tf(x_i)-v^Ty_i)^2\right\}+\lambda\|f\|^2\right\},
\end{equation}
where $p_{i}(f,v) = v^Tf(x_i)$, and $l_i(v) = v^Ty_i$ (see Appendix \ref{app: ridge}).

\textbf{Finite element methods:} Letting $\mathds{H}$ be an RKHS of functions satisfying some boundary conditions, and $V$ a finite-dimensional Hilbert space of functions, we minimize:
\begin{equation}
\min_{u\in\mathds{H}}\left\{\sup_{v\in V,\|v\|=1}\left\{(a(u,v) - f(v))^2\right\}\right\},
\end{equation}
where $a$ is a bilinear form and $f$ is a linear functional of $v$. For finite element methods, we typically do not add regularization. Instead, we require $a$ to be coercive to guarantee a stable solution, and we may sometimes use many different approaches to handle the boundary constraints. Here we assume we optimize over a space of functions that satisfy the boundary constraints. We see by adding the regularization term as in our theorem we guarantee a stable solution to the minimization problem without requiring $a$ to be coercive \cite{numerical}, Chapter 3.

\textbf{Learning PDEs:} In addition to solving PDEs, we may use our methods to learn PDEs. For example consider 
\begin{equation}
  u_t = f(u,u_x,u_{xx},...).  
\end{equation}
Defining $u_i = u(t,x_i)$ and $u_{i,x} = \partial_x u(t,x_i)$, we may solve this problem by minimizing
\begin{equation}\label{eq:PDE_loss}
\min_{f\in\mathds{H}}\left\{\sum_{i=1}^{N}\sup_{v\in V,\|v\|=1}\left\{\left(\int_{t_0}^{t_f}f(u_i,u_{i,x},...)v(t,x_i)dt - \int_{t_0}^{t_f}u_{i,t} v(t,x_i)dt\right)^2\right\}+\lambda\|f\|^2\right\}
\end{equation}
If $v(t,x_i)$ are chosen to be piecewise constant with support on the intervals $[t_j,t_{j+1}]$ we recover the loss in Section \ref{sec:pde}. Here $p_i(f,v)=\int f(u_i,u_{i,x},...)v(t,x_i)dt$ is the bilinear functional and $l_i(v)=\int u_{i,t} v(t,x_i)dt$ is the linear functional, see Appendix \ref{PDE_supp}. 

\textbf{Liouville operator occupation kernel method:} 
In \cite{OCKNew}, the Liouville operator method presented may be shown to be equivalent to minimizing:
\begin{equation}
\label{eq:luivOCK_2}
\inf_{u\in U}\left\{\sup_{v\in \mathds{V}, \|v\|=1}\left\{\left(\int_{0}^T \nabla\left(v(\gamma(t))\right)^Tf(\gamma(t))dt - \int_{0}^T\nabla\left(v(\gamma(t))\right)^Tu(\gamma(t))dt\right)^2\right\}\right\}
\end{equation}
where the bounded linear functional $l(v)$ is given by $ l(v) = \int_{0}^T \nabla\left(v(\gamma(t))\right)^Tf(\gamma(t))dt$ 
 and the bounded bilinear functional $p(v,u)$ is given by $p(v,u) = \int_{0}^T\nabla\left(v(\gamma(t))\right)^Tu(\gamma(t))dt.$
 
\textbf{Multivariate occupation kernel method (MOCK):} In this approach the authors minimize:
\begin{equation}
\min_{f\in \mathds{H}}\left\{\sum_{i=1}^{n}\sum_{j=1}^{n_i-1}\left\|\int_{t^i_j}^{t^i_{j+1}}\left(f(x^i(t)) - \dot{x}^i(t)\right)dt\right\|^2 + \lambda\|f\|^2_{\mathds{H}}\right\}
\end{equation}
which may be written under the form \eqref{eq:Loss} using the same techniques used to transform the regression problem \cite{MOCK}. In fact, MOCK is a regression problem with a linear functional for ``measuring'' $f$ that involves integration rather than the standard evaluation linear functional.

Still other applications include learning the drift and the diffusion of a stochastic differential equation. We now focus on two specific applications, namely ODEs in Section \ref{sec:ODE} and PDEs in Section \ref{sec:pde}. 
 
\section{ROCKing ODEs}
\label{sec:ODE}
Consider the problem of learning ODEs (or dynamical systems) from snapshots of trajectories (as presented in \cite{rielly2023learning}). We seek $f(x(t))$ that satisfies
\begin{equation}
    \dot{x}^i(t) = f(x^i(t)), \forall i \in [1,...,n]
\end{equation} Suppose we have $n$ trajectories, $x^1(t),...,x^n(t)\in \mathds{R}^d$, and for each $i\in 1,...,n$ there are $n_i$ samples of the trajectories. We minimize
\begin{multline}
\label{eq:WeakOCK}
\inf_{f\in\mathds{H}}\left\{\sum_{i=1}^n \sup_{v\in \mathds{V}_{i},\|v\|=1}\left\{\left(\int_{a^i}^{b^i}f(x^i(t))^Tv(t)dt - \int_{a^i}^{b^i}\dot{x}^i(t)^Tv(t)dt\right)^2\right\}+\lambda \|f\|^2_{\mathds{H}}\right\}
\end{multline}
It is worth noting that this generalizes the loss in \cite{MOCK} in that for an appropriate choice of quadrature and choice of test function spaces we recover MOCK.
Proceeding from (\ref{eq:WeakOCK}) and defining the kernel for $\mathds{H}$ to be $K_{\mathds{H}}$ and the kernel for $\mathds{V}_{i}$ to be $K_{\mathds{V}_i}(\cdot,*)=\Psi_i^T(\cdot)\Psi_i(*)$ with $\Psi_i(\cdot)\in \mathds{R}^{q\times d}$, we can apply Theorem \ref{thm:BigThm_sec} to get
\begin{equation}
    f(*) = \sum_{i=1}^n\left\{\int_{a^i}^{b^{i}}K_{\mathds{H}}(*,x^i(t))\left(\Psi_{i}^T(t)\right)dt\right\}\overrightarrow{\alpha}_{i}. \\
\end{equation}
Theorem \ref{thm:BigThm_sec} also tells us that we may obtain $\overrightarrow{\alpha_{i}}$ by solving \eqref{eq:linear_system}
with 
\begin{equation}
y_{i,k} = \int_{a^i}^{b^i}\dot{x}^i(t)^T\Psi_{i}(t)^Te_kdt
\mbox{ so that }
\overrightarrow{y}_{i}= \int_{a^i}^{b^i}\Psi_{i}(t)\dot{x}^i(t)dt
\end{equation}
and
\begin{equation}
M_{i,k:i',k'} =\int^{b^i}_{a^i}\int_{a^{i'}}^{b^{i'}}e_{k'}^T\Psi_{i'}(\tau)K_{\mathds{H}}(x^{i'}(\tau),x^{i}(t))\Psi_{i}(t)^Te_kd\tau dt
\end{equation}
so that the $M_{i:i'}$th block is 
\begin{equation}
\int^{b^i}_{a^i}\int_{a^{i'}}^{b^{i'}}\Psi_{i'}(\tau)K_{\mathds{H}}(x^{i'}(\tau),x^{i}(t))\Psi_{i}(t)^Td\tau dt.
\end{equation}
In our implementation, we choose 
\begin{equation}
\label{eq:KernelH}
K_{\mathds{H}}(\cdot,*) = k(\cdot,*)\otimes I_{d}
\mbox{ and }
\Psi_{i}(t) = \overrightarrow{\psi}_{i}(t) \otimes I_{d} \in \mathds{R}^{pd \times d}
\end{equation}
Where $\otimes$ denotes the Kronecker product. This choice of $\Psi$ is equivalent to using
\begin{equation}
\label{eq:KernelV}
\Psi_{i}(\cdot)^T\Psi_{i}(*) = \psi^T_{i}(\cdot)\psi_{i}(*)\otimes I_d
\end{equation}
%This is a seperable kernel proportional to the identity using the finite dimensional explicit kernel $\psi_{i}(\cdot)^T\psi_{i}(*)$. So for a scalar kernel with $p$ features, the vvRKHS of test functions we use is $pd$ dimensional. 
See \cite{MOCK} and Appendix \ref{app:Vectorizations} for a detailed analysis of the benefits of using kernels with this structure.
\subsection{Practical implementation}
While the derivations for ROCK methods can sometimes be involved, the implementation, even for very general methods allowing for flexibility in choices of quadratures and test function spaces, is often straightforward. Our implementation is 47 lines of Python code relying only on linear algebra operations (see Appendix \ref{app: implementation}). We accomplish this by applying a number of vectorizations, optimizations, and simplifications of the loss function for the case of learning dynamical systems.
By using the kernels in \eqref{eq:KernelH} and \eqref{eq:KernelV}, the linear system \eqref{eq:linear_system} becomes
\begin{equation}
\left(G\otimes I + \lambda I\right)\alpha = y
\end{equation}
where
\begin{multline}
G_{i,j} = \int_{a^i}^{b^i}\int_{a^j}^{b^j}k(x^i(t),x^j(s))\psi_i(t)\psi_j(s)^Tdt,ds \\ \approx \sum_{k=1}^{m}\sum_{l=1}^{m}q_kq_l k(x^i(t_k),x^j(t_l))\psi_i(t_k)\psi_j(t_l)^T\in \mathds{R}^{p\times p}.
\end{multline}
Here, $q_k$ and $q_l$ are the weights of the quadratures of integration, deriving for example from the trapezoid rule. Letting the matrix $A$ be such that stacking the columns of $A$ gives the vector $\alpha$, and defining $Y$ similarly, we reduce the  $dpn \times dpn$ linear system to the $pn \times pn$ linear system
\begin{equation}
\left(G+\lambda I\right)A = Y.
\end{equation} 
$d$ is the dimensionality of the system, $n$ the number of trajectories, and $p$ the number of test function features (see Appendix \ref{app:Computational}).

\section{ROCKing PDEs} \label{sec:pde}
Sometimes, a dynamical system is better interpreted as a PDE. A specific case of this is when the system is continuous in both space and time. We develop a version of ROCK to learn PDEs. Suppose, given data for $u(t,x)$, we wish to learn the following order-$d$ PDE:
\begin{equation} \label{pde_def}
    \partial_t u = f(u, \partial_x u, \partial_{xx} u, \dots).
\end{equation}
Assume the time domain $[0,T]$ is discretized into time points $t_j$ for $1\leq j \leq M$, and the spatial domain is discretized into points $x_i$ for $1 \leq i \leq N$. Let $u_i = u(\cdot, x_i)$. To learn $f$, we minimize the following cost function (which is a special case of \eqref{eq:PDE_loss}, see Appendix \ref{PDE_supp}):
\begin{equation}\label{pde_cost_1}
    J(f) = \sum_{j=1}^{M-1} \sum_{i=1}^N \left(u(t_{j+1}, x_i) - u(t_j, x_i) - \int_{t_j}^{t_{j+1}}f(u_i, \partial_x u_i, \partial_{xx} u_i, \dots) dt \right)^2 + \lambda \|f\|_{\mathds{H}}.
\end{equation}
$\mathds{H}$ is an RKHS with an explicit kernel with a feature map $\phi: \mathds{R}^{d+1} \rightarrow \mathds{R}^q$. If $f \in \mathds{H}$ then there exists an $\alpha \in \mathds{R}^q$ such that $f(u, \partial_x u, \partial_{xx} u, \dots) = \alpha^T \phi(u, \partial_x u, \partial_{xx} u, \dots)$ and the cost function becomes:
\begin{equation}\label{pde_cost_2}
    J(\alpha) = \| y - \Phi^T \alpha\|^2 + \lambda \alpha^T \alpha,
\end{equation}
where $y \in \mathds{R}^{N(M-1)}$. $y_{ij}$ is given by $u(t_j, x_i) - u(t_{j+1}, x_i)$ for each $i$ and $j$. $\Phi \in \mathds{R}^{q \times N(M-1)}$ is such that $\Phi_{ij} \in \mathds{R}^q$ is given by $\int_{t_j}^{t_{j+1}} \phi(u_i, \partial_x u_i, \partial_{xx} u_i, \dots) dt$. The minimizer of \eqref{pde_cost_2} is given by the solution to $(\Phi \Phi^T + \lambda I) \alpha = \Phi y$. 

If we are only given data for $u$, we must numerically compute spatial derivatives. In the experiments presented, we use finite differences to estimate the spatial derivatives. A coarsened spatial mesh is used to smooth errors in derivative computation. The explicit Euler method is used to integrate the PDE to generate new predictions. 
%We use the PDE model to forecast the Kuramoto-Shivashinsky (KS) system. A heatmap of the KS system is given in Figure~\ref{fig:dataset_fig}. Table~\ref{table:Task1Results} includes results for the PDE model under the "$KS^*$" columns. 
%While the PDE model outperforms the competition for E1 scores, we see our results for E2 are rather unsatisfying. This indicates the PDE approach requires more work. Another limitation of the PDE model is choice of integrator. Even if the the model succeeds in learning the underlying PDE, forecasts can be poor if the wrong integrator is used and the PDE is sufficiently stiff.

\section{Data and experiments}
\begin{table}[h]
\caption{Summary of MOCK and CTF challenge's datasets}
\centering
\begin{minipage}[t]{0.48\textwidth}
\centering
\subcaption{MOCK datasets}
\begin{tabular}{rrrr}
\hline
Dataset & $d$ & $n$ & $m$ \\
\hline
NFHN & 2 & 150 & 201 \\
Lorenz63 & 3 & 150 & 201 \\
Lorenz96-16 & 16 & 100 & 81 \\
Lorenz96-32 & 32 & 100 & 81 \\
Lorenz96-128 & 128 & 100 & 81 \\
Lorenz96-1024 & 1024 & 100 & 100 \\
\hline
\end{tabular} \label{MOCKdata}
\end{minipage}%
\hfill
\begin{minipage}[t]{0.48\textwidth}
\centering
\subcaption{CTF datasets}
\begin{tabular}{rrrr}
\hline
Dataset & $d$ & $n$ & $m$ \\
\hline
Double pendulum & 4 & 1 & 2000 \\
Lorenz & 3 & 1 & 2000 \\
Rossler & 3 & 1 & 2000 \\
Kolmogorov & 16384 & 1 & 2000 \\
Kuramoto-Shivashinsky & 1024 & 1 & 1000 \\
\hline
\end{tabular} \label{competitiondata}
\end{minipage}
% \vspace{1em}
\begin{flushleft}
\footnotesize
    $d$ denotes the dimension of the dynamical system, $n$ the number of trajectories, and $m$ the number of samples per trajectory.
\end{flushleft}
\end{table}

\subsection{Benchmarking against published methods}
\label{sec:Experiments}
As we generalize MOCK with ROCK, we benchmark against published methods from \cite{rielly2023learning}, comparing on the same datasets with the same training procedures. These methods include:\\
\textbf{Sparse Identification of Nonlinear Dynamics (SINDy):} SINDy can be thought of as a solution for learning dynamical systems that employs sparsity promoting regression \cite{brunton2016discovering}\cite{kaheman2020sindy, fasel2022ensemble}.\\
\textbf{Extended Dynamic Mode Decomposition (eDMD) \cite{tu2013dynamic}, \cite{williams2015data}} We benchmark against eDMD-RFF \cite{Lew2023}, eDMD-Poly \cite{williams2015data}, and eDMD-Deep \cite{yeung2019learning}.\\
\textbf{Deep Learning Methods (ResNet)} See \cite{lusch2018deep, yeung2019learning, li2019learning}.
Deep learning methods can be incorporated into the SINDy framework to identify sparse, interpretable, and predictive models from data \cite{champion2019data, bakarji2022discovering}. We benchmark with ResNet \cite{he2016deep} \cite{lu2021deepxde}.\\
\textbf{Latent ODEs for Irregularly-Sampled Time Series (LODE)}
The Latent ODE method (also a deep learning method \cite{rubanova2019latent}) is an update of the Neural ODEs model introduced in \cite{chen2018neural}.\\
\textbf{Multivariate Occupation Kernel Method (MOCK):} ROCK generalizes this method \cite{rielly2023learning}.

\subsubsection{Datasets}\label{sec:datasets}

 %We test the methods on a diverse set of synthetic and real-world datasets that reflect the challenges of learning systems of ODEs. A summary of the datasets is provided in table \ref{table:datasets} in appendix \ref{App: Datasets}.
 
 We test the methods on a subset of the synthetic datasets in \cite{rielly2023learning}. These include datasets with and without noise, going up to 1024 dimensions.\\
\textbf{Noisy Fitz-Hugh Nagumo (NFHN)} The FitzHugh-Nagumo oscillator 
\cite{fitzhugh1961impulses} is a nonlinear 2D dynamical system that models the basic behavior of excitable cells, such as neurons and cardiac cells \cite{rielly2023learning}. Considerable noise was added. See Figure 3 in \cite{rielly2023learning}. \\
\textbf{Noisy Lorenz63 (Lorenz63)} The Lorenz63 system \cite{lorenz1963deterministic} is a 3D system provided as a simplified model of atmospheric convection. This dataset also contains added noise.\\
\textbf{Lorenz96}
The Lorenz96 data arises from \cite{75462} in which a system of equations is proposed that may be chosen to have any dimension greater than 3. (Lorenz96-16) has 16 dimensions, etc. The synthetic data we benchmark against includes Lorenz96 of dimension 16, 32, 128 and 1024. 

A summary of the mentioned datasets is reported in Table \ref{MOCKdata}
\subsubsection{Experimental methodology}

For each of the datasets, we use implicit radial basis kernels to define our vvRKHS for the model: Gaussian ($g$), Laplace ($l$), and the $C^{10}$ Mat\'ern ($m$) defined by
\begin{equation}
g(r) = e^{-\frac{r^2}{2\sigma^2}}
,  \quad
l(r) = e^{-\frac{r}{\gamma}}, \quad \text{and}
\end{equation}
\begin{equation}
m(r)  = \frac{e^{-\frac{r}{\gamma}}}{945} \left(\left(\frac{r}{\gamma}\right)^5 + 15\left(\frac{r}{\gamma}\right)^4\right. + \left.105\left(\frac{r}{\gamma}\right)^3 + 420\left(\frac{r}{\gamma}\right)^2 + 945 \frac{r}{\gamma} + 945\right) 
\end{equation}
respectively, where $r = \|x-y\|$ and $\gamma$ and $\sigma$ are the scale parameters. 

We use Legendre polynomials (see Appendix \ref{app:legendre}) for our space of test functions allowing the number of basis functions to be a tunable hyperparameter. 
\begin{comment}
We split the dataset into 60\% training, 20\% primary validation and 20\% secondary validation to validate for our choice of kernel, scale parameter, regularization parameter, number of test features in our test function space, and trajectories' lengths. We used the primary validation set to select the scale and regularization parameters and the secondary validation set to select the kernel, the number of test features, and the trajectories' lengths.  
We then retrained on the training and 20\% primary validation set, validating on the 20\% secondary validation set for the scale and regularization parameter. This is the final model we integrated on a disjoint test set to publish our errors. We publish the errors defined in \cite{rielly2023learning}, reporting root mean squared error (RMSE) on the full test trajectories as Err and the average RMSE on the next sample as 1-ERR (see Table \ref{table:Results}).
\end{comment}

We split the dataset into three subsets: 60\% for training, 20\% for primary validation, and 20\% for secondary validation. The primary validation set was used to tune the scale and regularization parameters. The secondary validation set was used to select the kernel type, the number of test features in the test function space, and the trajectory lengths.

After selecting these hyperparameters, we retrained the model on the combined training and primary validation sets (80\%), and used the secondary validation set (20\%) to re-tune the scale and regularization parameters. This final model was then used to integrate (predict) on a separate, disjoint test set.

We report prediction errors as defined in \cite{rielly2023learning}: the root mean squared error (RMSE) over the full test trajectories, denoted as Err, and the average RMSE for the next-step prediction, denoted as 1-ERR (see Table 
 \ref{table:Results}). 
\subsubsection{Performances}
The ROCK algorithm greatly reduces the error on the NFHN and Lorenz63 datasets. These two datasets have significant noise, indicating that ROCK generalises to noisy data better than MOCK. ROCK has excellent performances overall. However, the LODE method remains a serious challenger for Err in high dimensions.   

\begin{table}[H]
\centering
\caption{Dynamical system estimation results}\label{table:Results}
\vspace{1\baselineskip}
\begin{adjustbox}{max width=\textwidth}
\begin{tabular}{l
                *{6}{r@{\hskip 3pt}r}} % 6 datasets x 2 subcols (Err, 1-Err)
\toprule
\multirow{2}{*}{Method} 
& \multicolumn{2}{c}{Lorenz63} 
& \multicolumn{2}{c}{NFHN} 
& \multicolumn{2}{c}{Lorenz96-16} 
& \multicolumn{2}{c}{Lorenz96-32} 
& \multicolumn{2}{c}{Lorenz96-128} 
& \multicolumn{2}{c}{Lorenz96-1024} \\
\cmidrule(r){2-3} \cmidrule(r){4-5} \cmidrule(r){6-7} 
\cmidrule(r){8-9} \cmidrule(r){10-11} \cmidrule(r){12-13}
& Err & 1-Err 
& Err & 1-Err 
& Err & 1-Err 
& Err & 1-Err 
& Err & 1-Err 
& Err & 1-Err \\
\midrule
ResNet 
& 2.07 & .014 
& 5.94 & .064 
& 6.42 & .010 
& 11.56 & .020 
& 24 & 2.64 
& 66.1 & 6.48 \\
eDMD 
& 1.93 & .009 
& 4.01 & .036 
& 5.93 & .029 
& 8.10 & .028 
& 16 & .089 
& 26.9 & .38 \\
SINDy-Poly 
& .99 & .002 
& 1.68 & .022 
& .40 & .0003 
& 8.56 & .037 
& 19 & .066 
& NA & NA \\
MOCK 
& .65 & .002 
& 1.11 & .028 
& .22 & .0001 
& .47 & .0001 
& 16 & .035 
& 17.8 & \textbf{.013} \\
ROCK 
& \textbf{.51} & \textbf{.001} 
& \textbf{.41} & \textbf{.012} 
& \textbf{.13} & \textbf{.0001} 
& \textbf{.41} & \textbf{.0001} 
& \textbf{15} & \textbf{.019} 
& 17.0 & .014 \\
LODE 
& 1.49 & .092 
& .64 & .131 
& 3.82 & .138 
& 6.06 & .217 
& \textbf{15} & .342 
& \textbf{15.8} & 1.04 \\
\bottomrule
\end{tabular}
\end{adjustbox}

\begin{flushleft}
\footnotesize
Minimum (best) values are in bold. MOCK (best of Gaussian, Laplace, Mat\'ern, and random Fourier features), ROCK, SINDy-Poly, eDMD (best of -Deep, -Poly, -RFF), ResNet, and Latent ODE (LODE) are compared. No result could be obtained for SINDy-Poly on Lorenz96-1024. See Section \ref{sec:datasets} for details of the datasets. All experiments were run on google collab without use of GPUs.
\end{flushleft}
\end{table}

\subsection{Benchmarking against the Common Task Framework challenge}

In addition to the experiments in Section \ref{sec:Experiments}, we benchmark ROCK on the datasets provided by the  Common Task Framework (CTF) of the AI Institute for Dynamical Systems \cite{M4DL}.%\href{https://maths4dl.ac.uk/wp-content/uploads/2024/04/Kutz\_CTF.pdf}{[1]}. 
\subsubsection{CTF datasets}
\textbf{Double Pendulum (DP):} This dataset provides angle and angular momentum measurements of the dynamics derived from a double pendulum. \\
\textbf{Lorenz63 (LR):} This dataset provides evenly sampled location from the Lorenz63 dynamical system.\\
\textbf{Rossler (RS):} This is also a three-dimensional dynamical system designed to have similar chaotic behavior as the Lorenz attractor but be easier to analyze.\\
\textbf{Kolmogorov (KMG):} KMG arises from a nonlinear time-dependent partial differential equation of two spatial variables. \\
\textbf{Kuramoto-Shivashinsky (KS):} KS arises from a 4th order non-linear PDE. The PDE is time-dependent and defined on a one-dimensional spatial domain. It is highly sensitive to initial conditions and exhibits chaotic behavior. 

We benchmark on Task 1 for which a noiseless trajectory of length $m$ is provided as a training set, and the goal is to predict the next $m$ time-points. A summary of the CTF Task 1 datasets is reported in Table \ref{competitiondata}. 
\begin{figure}[h]
\centering

\begin{subfigure}[t]{0.19\textwidth}
    \includegraphics[width=\linewidth]{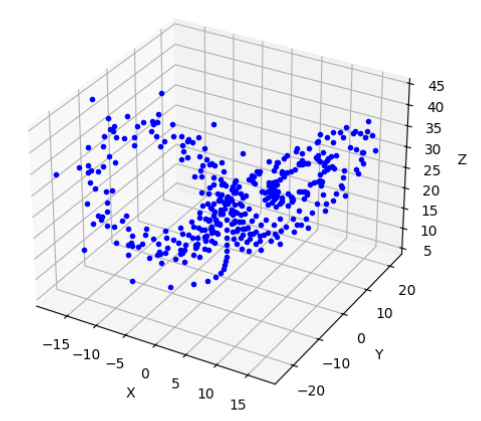}
    \caption{LR}
\end{subfigure}
\hfill
\begin{subfigure}[t]{0.19\textwidth}
    \includegraphics[width=\linewidth]{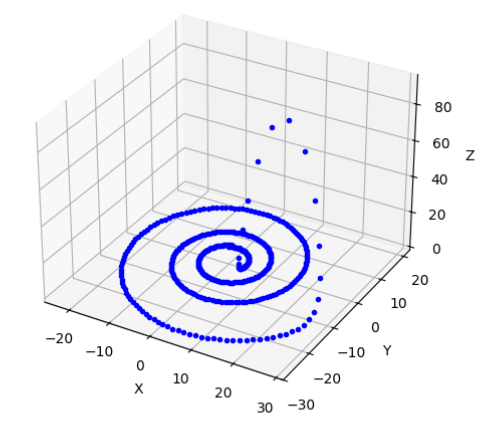}
    \caption{RS}
\end{subfigure}
\hfill
\begin{subfigure}[t]{0.19\textwidth}
    \includegraphics[width=\linewidth]{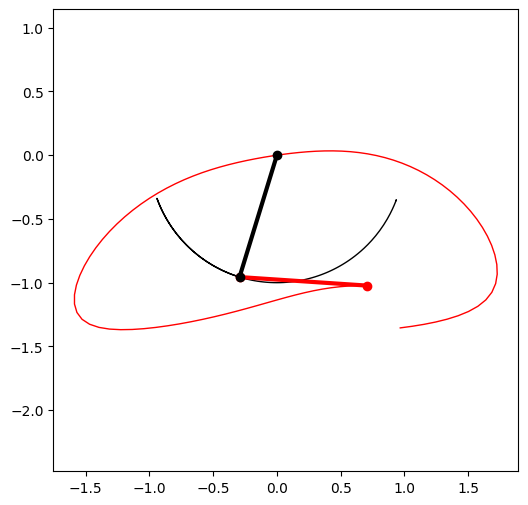}
    \caption{DP}
\end{subfigure}
\hfill
\begin{subfigure}[t]{0.19\textwidth}
    \includegraphics[width=\linewidth]{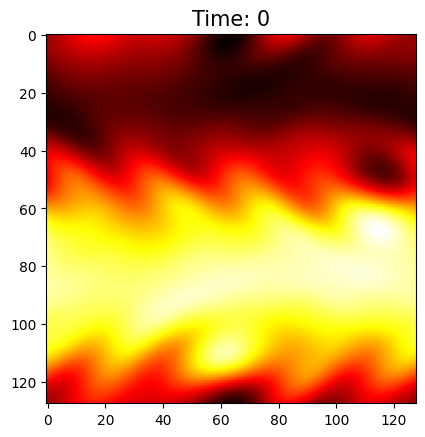}
    \caption{KMG}
\end{subfigure}
\hfill
\begin{subfigure}[t]{0.19\textwidth}
    \includegraphics[width=\linewidth]{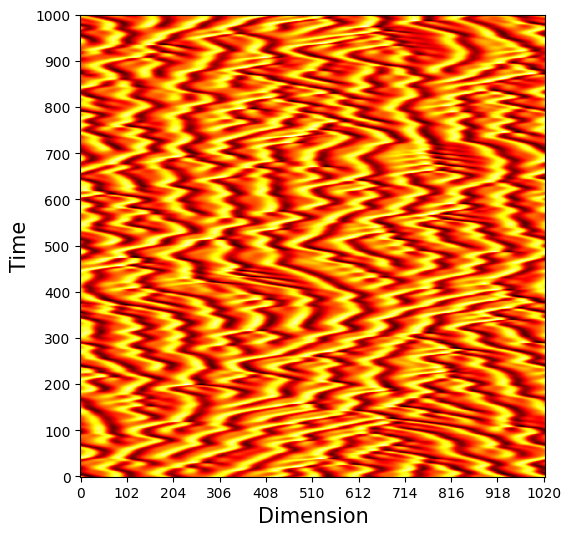}
    \caption{KS}
\end{subfigure}
\caption{Dynamic and spatio-temporal systems from CTF challenge}\label{fig:dataset_fig}
\end{figure}
\subsubsection{Experiments for CTF}
The hyperparameters for the ROCK model for ODEs are the number of test features, the number of training and validation trajectories as well as the choice of kernel, scale and regularization parameters.
We apply the same validation procedure as described in Section \ref{sec:Experiments}.

%\begin{enumerate}
%    \item The number of test features 
%    \item The number of training trajectories to cut the training set into
%    \item The number of validation trajectories to cut the validation set into.
%    \item The choice of kernel
%    \item The scale parameter for the kernel.
%    \item the regularization term for the model.
%\end{enumerate}
The choice of test function space and trajectory lengths determines the number of parameters of the learned model. The number of parameters increases linearly with the number of trajectories, the number of test features and the number of dimensions. Thus cutting the trajectories into shorter trajectories increases the model size. This is demonstrated in Table \ref{table:numparams} Appendix \ref{app:comp results}. 

The KS system arises from a PDE and requires a simplified validation procedure, optimizing only on the scale and regularization parameters using 80\% training and 20\% validation sets. This model has one-dimensional test features, and Gaussian random Fourier features with periodic boundary conditions for the kernel. The space derivatives were estimated using central finite differences on a coarser grid to provide more stable estimates.   

We report the $E1$ and $E2$ scores provided to us from the platform. $E1$ measures our model's ability to do short-term predictions, while $E2$ measures our model's ability to do long-term predictions. These scores range from 100\% (perfect) to $-\infty$ with 0\% indicating a trivial result, see \cite{M4DL}.
%\href{https://maths4dl.ac.uk/wp-content/uploads/2024/04/Kutz\_CTF.pdf}{[1]}.
\subsubsection{Experimental results}

MOCK and ROCK models currently outperform all other submissions to the competition on both metrics for the first task, with the exception of KS, E2.  However, the competition is still ongoing, so we expect many more entries to arrive. In Table \ref{table:Task1Results}, we compare the performance of our best-performing ROCK submission to the best-performing MOCK submission. While the E1 scores tend to be similar for both types of models, the E2 scores can be much better (by as much as 23\%) with the ROCK models. Although all datasets here are noise free, ROCK models appear to generalize better. ROCK models tend to have far fewer parameters (see the discussion in Appendix \ref{app:comp results}), which may explain this superior generalization. In addition, ROCK outperforms MOCK in 5 of the 8 ODE cases. Proper and complete hyper-parameter tuning of ROCK should always ensure ROCK at least matches MOCK. We included the current best results (by May 14th, 2025) from other competition participants on the leader board as a point of comparison.

The KS results (emphasized in Table \ref{table:Task1Results} with *) apply ROCK to learning PDEs, as discussed in Section \ref{sec:pde}.  We were unable to learn KS with an ODE model. However, we were able to learn a reasonable solution by treating the problem as a PDE. This indicates there are datasets that benefit from a treatment of the dynamics in a PDE context. We reported the best score for E1 on KS. However, our scores for the long-term E2 prediction still need improvement.  

We report the model sizes for MOCK and ROCK. We see ROCK can use 10 to 20 times fewer parameters while still matching or outperforming MOCK. The hyperparameters of ROCK may be chosen specifically to provide models with very low parameter counts (see Table \ref{table:numparams} in Appendix \ref{app:comp results}).
\begin{table}[h]
\caption{CTF leader board results}
\begin{small}
\begin{center}
\begin{tabular}{cccccc}
\toprule E1& DP & KMG & LR& RS &  KS$^*$ \\
\midrule
MOCK& \textbf{97.27} & \textbf{90.71} & 97.24& 99.98&  - \\
ROCK& 96.63 & 90.11 & \textbf{99.1}& \textbf{99.98}& \textbf{93.77}\\
Best of others& 95.94 & - & 78.11
& 95.30 & - 29.52\\\midrule E2 & DP& KMG & LR & RS & KS$^*$ \\\midrule MOCK & 83.70& 88.18 & 60.2 & \textbf{87.40}& -\\ ROCK & \textbf{94.78}& \textbf{94.81} & \textbf{83.95} & 85.97 & 33.01 \\ Best of others&77.78& - & 82.63 & 63.68
& \textbf{92.72 }
\\\midrule Size & DP& KMG & LR & RS & KS$^*$  \\ \midrule MOCK & 4k& \textbf{16.4M} & 3k & 3k& -\\ ROCK & \textbf{480}& 49.2M & \textbf{1k} & \textbf{160}&\textbf{200} \\\bottomrule
\end{tabular}
\end{center}
\end{small}
\begin{flushleft}
\footnotesize
E1 and E2 are metrics for measuring performance. See \cite{M4DL}. %\href{https://maths4dl.ac.uk/wp-content/uploads/2024/04/Kutz\_CTF.pdf}{[1]}. 
 DP: double pendulum. KMG: Kolmogorov, LR: Lorenz63, RS: Rossler, KS: Kuramoto-Shivashinsky. Size: \# of parameters (k: thousand, M: million). All experiments require no more than 20GB of RAM and do not require GPU.
\end{flushleft}
\label{table:Task1Results}
\end{table}

\section{Discussion}
%We see ROCK for learning ODEs (which generalizes MOCK) either performs up to par with MOCK or outperforms MOCK in all cases. We see ROCK typically performs better on the long term forecasing tasks indicating that ROCK generalizes better. This is not surprising considering the total number of parameters for ROCK models are often orders of magnitude smaller than for MOCK models. That is, ROCK allows experimenters to control the model sparsity of the final model. Finally, we explore some simple modifications to ROCK (by adding a regression preprocessing step, using compactly supported kernels, and by using ROCK to learn PDEs) and observe additional improvements to performance and computational complexity. These experiments suggest this is a powerful and very general optimization framework with wide reaching future applications.
%%%%%%%%%%%%%%%%%%%%%%%%%%%%%%%%%%%%%%%%%%%%%%%%%%%%%%%%%%%%
In this work, we introduced ROCK, a unified variational framework for a broad class of estimation problems, including regression and dynamical systems modeling. Our formulation generalizes MOCK, an earlier approach that was proposed for learning dynamical systems. It provides a principled view of vector field estimation, hence trajectory forecasting, within a tractable optimization setting grounded in vvRKHSs.

Through experiments on both synthetic datasets and the common task framework, we find that ROCK consistently performs on par with other state-of-the-art methods, including MOCK. This is especially notable given that ROCK models often have orders of magnitude fewer parameters than their counterparts. The ability to control model complexity through kernel design and regularization gives ROCK a distinct advantage in scenarios where interpretability, sparsity, or data efficiency is critical.

In future work, we aim to extend the framework to broader classes of PDEs, including systems with various boundary conditions. Another promising direction is to treat trajectories as latent variables and develop an iterative algorithm that alternates between estimating trajectories and updating the vector field. We hypothesize that this approach will be beneficial in the case of very sparse observations of the trajectories for ODEs. This can also be used to help estimating partial derivatives for PDEs.

\newpage

\printbibliography
\newpage
\appendix

%\section{Appendix / supplemental material}
%\subsection{Results of the compactly supported kernel experiments}
%We define the density of an $M\times N$ sparse matrix to be the number of non-zero entries in the matrix divided by $MN$. To compare the two proposed %methods in section \ref{sec:sparse}, we subsample the training set for use with the Mat\'ern kernel so that the number of nonzero entries in the Gram matrix %is the same as that of the CSWK for a given density. The errors are computed using (28) in \cite{MOCK} and the results are then plotted together. See figure %\ref{fig:cswk}.
%\begin{figure}[h]\centering
%\includegraphics[width=0.5\textwidth]{density_vs_error.png}
%\caption{A comparison of the error using two methods to handle large data sets. The blue circles indicate the results of the compactly supported kernel %experiments, while the red plus signs show results for the Mat\'ern kernel. We see that compactly supported kernels yield better errors on the test set in %comparison to naively subsampling the training data and using a kernel with global support. } \label{fig:cswk}
%\end{figure}
\section{vvRKHSs} \label{app: vvrkhs}
If $\mathds{H}$ is a vvRKHS we may observe the following. 
\begin{enumerate}
\item For every $l_{v,x}$ Riesz representation theorem guarantees the existence of a unique $l^*_{v,x}\in \mathds{H}$ for which
\begin{equation}
l_{v,x}(f) = f(x)^Tv = \left<f,l^*_{v,x}\right>_{\mathds{H}}
\end{equation}
Observe that for any $v,w\in \mathds{R}^d$ and $x,y\in \mathcal{X}$
\begin{align}
l_{v,x}(l^*_{w,y})&= l^*_{w,y}(x)^Tv \\ 
&= \left<l^*_{w,y},l^*_{v,x}\right>_{\mathds{H}} \\
&= \left<l^*_{v,x},l^*_{w,y}\right>_{\mathds{H}}\\
&= l_{w,y}(l^*_{v,x}) \\
&= l^*_{v,x}(y)^Tw
\end{align}
Thus $l_{v,x}(l^*_{w,y})$ is a bilinear functional of $v$ and $w$. It can therefor be written in the form
\begin{equation}
v^TK(x,y)w
\end{equation}
for some matrix valued $K(x,y)$. This in turn implies
\begin{equation}
l^*_{v,x}(y)^Tw = v^TK(x,y)w \Rightarrow l^*_{v,x}(y)^T = v^TK(x,y)
\end{equation}
and similarly
\begin{equation}
l^*_{w,y}(x)^T = w^TK^T(x,y)
\end{equation}
but since we chose $w,v,y,x$ arbitrarily we also have (after substituting $w$ for $v$ and $y$ for $x$)
\begin{equation}
l^*_{w,y}(x)^T = w^TK(y,x)
\end{equation}
Thus
\begin{equation}
K^T(x,y) = K(y,x)
\end{equation}

Using $v = e_i$ and $w = e_j$ we can define $K(x,y)$ by
\begin{equation}
e_i^TK(x,y)e_j = l_{e_i,x}(l^*_{e_j}(y)) = l_{e_j,y}(l^*_{e_i}(x)) = K(x,y)_{i,j}
\end{equation}
\item The kernel function $K(x,y)$ is positive semidefinite. In other words, for any $x_1,...,x_n\in \mathcal{X}$, $\alpha_1,...,\alpha_n\in \mathds{R}^d$,
\begin{equation}
\sum_{i=1}^{n}\sum_{j=1}^{n}\alpha_i^TK(x_i,x_j)\alpha_j \geq 0
\end{equation}
To see this observe:
\begin{align}
\sum_{i=1}^{n}\sum_{j=1}^{n}\alpha_i^TK(x_i,x_j)\alpha_j &= \sum_{i=1}^{n}\sum_{j=1}^{n}\left<l^*_{x_i,\alpha_i},l^*_{x_j,\alpha_j}\right> \\
&= \left<\sum_{i=1}^{n}l^*_{x_i,\alpha_i},\sum_{j=1}^{n}l^*_{x_j,\alpha_j}\right> \\
&= \left\|\sum_{i=1}^{n}l^*_{x_i,\alpha_i}\right\|^2_{\mathds{H}} \geq 0
\end{align}

\item We also have the reproducing property. Recalling 
\begin{equation}
l^*_{v,x}(y)^T = v^TK(x,y) \Rightarrow l^*_{v,x}(y) = K^T(x,y)v = K(y,x)v
\end{equation}
we have
\begin{equation}
f(x)^Tv = \left<f,l^*_{v,x}(\cdot)\right> = \left<f,K(\cdot,x)v\right>
\end{equation}
\end{enumerate}
All finite dimensional Hilbert spaces are equivalent (in the sense that their norms are equivalent). Therefore, all finite dimensional Hilbert spaces of functions are reproducing kernel Hilbert spaces. Here we demonstrate how to construct a kernel for any Hilbert space of vector valued functions. This very simple and well-known result is often useful so it is provided below for completeness.
\begin{proposition}
    \label{Thm:HspaceIsRKHS}
    Let $\mathds{H}$ be a Hilbert space of vector valued functions where for each $f, g \in \mathds{H}$, $f(x) \in \mathds{R}^d$ takes $x\in \mathcal{X}$ to $\mathds{R}^d$ with real valued inner product $\left<f,g\right>_{\mathds{H}} \in \mathds{R}$. And let $\{b_1(x),...,b_n(x)\}$ be an orthonormal basis for this Hilbert space of functions. Then $\mathds{H}$ is a vector valued RKHS with kernel function
    \begin{equation}
    K_{\mathds{H}}(x,y) = \Psi(x)^T\Psi(y)
    \end{equation}
    where 
    \begin{equation}
    \Psi(x) = \left[\begin{array}{c}
    b_1(x)^T \\
    \vdots \\
    b_n(x)^T\end{array}\right]
    \end{equation}
\end{proposition} 

\section{Proof of the representer theorem}
\label{app:thm1_2}
The following proposition will be helpful in the proof of Theorem \ref{thm:BigThm_sec}.
\begin{proposition}[Riesz Representation]
\label{thm:Riesz_2}
Let $\mathds{H}$ be a vector valued RKHS taking a set $\mathcal{X}$ to $\mathds{R}^d$, and $l$ be a \textbf{bounded} linear functional taking $\mathds{H}$ to $\mathds{R}$ then there is a unique $l^*\in \mathds{H}$ for which for any $h\in \mathds{H}$
\begin{displaymath}
l(h) = \left<h,l^*\right>_{\mathds{H}}
\end{displaymath}
Where $\left<\cdot,*\right>_{\mathds{H}}$ is the inner product in $\mathds{H}$.
Moreover $l^*$ is given by
\begin{displaymath}
l^*(x) = \sum_{i=1}^{d}l\left(K_{\mathds{H}}(\cdot,x)e_i\right)e_i
\end{displaymath}
Where $K_{\mathds{H}}(\cdot,*)$ is the reproducing kernel for the RKHS.
\end{proposition}
We include the proof of Proposition \ref{thm:Riesz_2} for completeness.
\begin{Proof}[Representer theorem]
Existence and uniqueness is guaranteed by the Riesz representation theorem. To derive the form of $l^*$ let $w\in \mathds{R}^d$ and $x \in \mathcal{X}$ be arbitrary and consider (using the reproducing property of vector valued RKHSs)
\begin{subequations}
\begin{align}
w^Tl^*(x) &= \left<l^*, K_{\mathds{H}}(\cdot,x)w\right>_{\mathds{H}} \\ 
&= \left<K_{\mathds{H}}(\cdot,x)w,l^*\right>_{\mathds{H}} \\
&= l(K_{\mathds{H}}(\cdot,x)w) \\
&= l\left(K_{\mathds{H}}(\cdot,x)I_dw\right) \\
&= l\left(K_{\mathds{H}}(\cdot,x)\left(\sum_{i=1}^{d}e_ie_i^T\right)w\right) \\
&= l\left(\sum_{i=1}^{d}K_{\mathds{H}}(\cdot,x)e_i(e_i^Tw)\right) \\
&= \sum_{i=1}^{d}l\left(K_{\mathds{H}}(\cdot,x)e_i\right)e_i^Tw \\
&= \sum_{i=1}^{d}w^Te_il\left(K_{\mathds{H}}(\cdot,x)e_i\right) \\
&= w^T\left(\sum_{i=1}^{d}l\left(K_{\mathds{H}}(\cdot,x)e_i\right)e_i\right)
\end{align}
\end{subequations}
Where we use the reproducing property, symmetry of real inner product, the definition of $l^*$, a clever form of multiplying by an identity, and linearity of the linear functional. Notice, in order to move the sum outside the linear functional, we also require $K_{\mathds{H}}(\cdot,x)e_i \in \mathds{H}$ which holds by the definition of vvRKHS's. Since $w$ and $x$ were chosen arbitrarily, we may conclude that 
\begin{displaymath}
l^*(x) = \sum_{i=1}^{d}l\left(K_{\mathds{H}}(\cdot,x)e_i\right)e_i
\end{displaymath}
\end{Proof}
We now give the proof Theorem \ref{thm:BigThm_sec}.
\begin{Proof}[Representer theorem]
We break this proof up into three separate parts. In part one, we simplify and evaluate the supremum, in part two we evaluate the inner products in terms of the linear functionals and bilinear forms, and in the final part, we demonstrate a representer theorem argument to arrive at our final conclusion.\\
\textbf{Part 1:}
Fix $f$ and consider 
\begin{equation}
\sup_{v\in\mathds{V}_i,\|v\|_{\mathds{V}_i}=1}\left\{\left(p_i(f,v)-l_i(v)\right)^2\right\}
\end{equation}
Since $p_i$ and $l_i$ are bounded, we can apply Proposition\ref{thm:Riesz_2} to rewrite the interior of the supremum as 
\begin{equation}
\left(p_i(f,v)-l_i(v)\right)^2 = \left(\left<v,\mathcal{L}^*_{i,f}\right>_{\mathds{V}_i}-\left<v,l^*_{i}\right>_{\mathds{V}_i}\right)^2
\end{equation}
with $\mathcal{L}^*_{i,f}\in\mathds{V}_i$ and $l^*_i\in\mathds{V}_i$ given by
\begin{equation}
\mathcal{L}^*_{i,f}(*) = \sum_{j=1}^{d_i}p_i(f(\cdot),K_{\mathds{V}_i}(\cdot,*)e_j)e_j 
\end{equation}
and
\begin{equation}
l^*_{i}(*) = \sum_{j=1}^{d_i}l_i\left(K_{\mathds{V}_i}(\cdot,*)e_j\right)e_j
\end{equation}
Now
\begin{align}
\sup_{v\in\mathds{V}_i,\|v\|_{\mathds{V}_i}=1}\left\{\left(\left<v,\mathcal{L}^*_{i,f}-l^*_i\right>_{\mathds{V}_i}\right)^2\right\}
&= \left(\left<\frac{\mathcal{L}^*_{i,f}-l^*_i}{\|\mathcal{L}^*_{i,f}-l^*_i\|_{\mathds{V}_i}},\mathcal{L}^*_{i,f}-l^*_i\right>_{\mathds{V}_i}\right)^2 \\
&= \frac{\|\mathcal{L}^*_{i,f}-l^*_i\|_{\mathds{V}_i}^4}{\|\mathcal{L}^*_{i,f}-l^*_i\|_{\mathds{V}_i}^2} \\
&= \|\mathcal{L}^*_{i,f}-l^*_i\|_{\mathds{V}_i}^2 \\
& = \left<\mathcal{L}^*_{i,f}-l^*_i,\mathcal{L}^*_{i,f}-l^*_i\right>_{\mathds{V}_i}
\end{align}
Therefore
\begin{equation}
(\ref{eq:Loss})\Leftrightarrow\inf_{f\in\mathds{H}}\left\{\sum_{i=1}^{n}\left(\left<\mathcal{L}^*_{i,f}-l^*_i,\mathcal{L}^*_{i,f}-l^*_{i}\right>_{\mathds{V}_i}\right)+\lambda(\|f\|^2_{\mathds{H}})\right\}
\end{equation}
\textbf{Part 2:}
Notice, our inner products are real valued so $\left<l^*_{i},\mathcal{L}^*_{i,f}\right>_{\mathds{V}_i} = \left<\mathcal{L}^*_{i,f},l^*_i\right>_{\mathds{V}_i}$. Lets evaluate
\begin{equation}
\left<\mathcal{L}^*_{i,f},\mathcal{L}^*_{i,f}\right>_{\mathds{V}_i}, \left<l^*_i,\mathcal{L}^*_{i,f}\right>_{\mathds{V}_i}, \left<l^*_i,l^*_i\right>_{\mathds{V}_i}
\end{equation}
\begin{align}
\left<l^*_i,l^*_i\right>_{\mathds{V}_i}&= l_i(l_i^*(*))\\
&= l_i\left(\sum_{j=1}^{d_i}l_i(K_{\mathds{V}_i}(\cdot,*)e_j)e_j\right) 
%&= \sum_{j=1}^{d_i}l_i\left(l_i(K_{\mathds{V}_i}(\cdot,*)e_j)e_j\right)
\end{align}
Where the inside linear functional is taken over $\cdot$ and the outside linear functional over $*$. Recall we define $d_i$ to be the dimensionality of the range of the functions in $\mathds{V}_i$. In addition, assuming $K_{\mathds{V}_i}$ is finite dimensional of dimension $q_i$ (i.e. with $q_i$ features), so that 
\begin{equation}
K_{\mathds{V}_i}(\cdot,*) = \Psi_i(\cdot)^T\Psi_i(*) \in \mathds{R}^{d_i\times d_i}
\end{equation}
for some 
\begin{equation}
\Psi_i:\mathcal{Y}_i \Rightarrow \mathds{R}^{q_i\times d_i}
\end{equation}
and assuming $e_k \in span\{\Psi_i(*)\alpha, \alpha \in \mathds{R}^{d_i},*\in \mathcal{X}\}$ for all $k$,
We may further simplify as follows:
\begin{align}
\left<l^*_i,l^*_i\right>_{\mathds{V}_i}&= l_i\left(\sum_{j=1}^{d_i}l_i(\Psi_i(\cdot)^T\Psi_i(*)e_j)e_j\right) \\
&= l_i\left(\sum_{j=1}^{d_i}l_i\left(\Psi_i(\cdot)^T\left(\sum_{k=1}^{q_i}e_ke_k^T\right)\Psi_i(*)e_j\right)e_j\right) \\
&= l_i\left(\sum_{j=1}^{d_i}\sum_{k=1}^{q_i}l_i\left(\Psi_i(\cdot)^T\left(e_ke_k^T\right)\Psi_i(*)e_j\right)e_j\right) \\
&= l_i\left(\sum_{j=1}^{d_i}\sum_{k=1}^{q_i}l_i\left(\Psi_i(\cdot)^Te_k\right)\left(e_k^T\Psi_i(*)e_j\right)e_j\right) \\
&= l_i\left(\sum_{j=1}^{d_i}\sum_{k=1}^{q_i}l_i\left(\Psi_i(\cdot)^Te_k\right)e_j\left(e_j^T\Psi_i(*)^Te_k\right)\right) \\
&= l_i\left(\sum_{k=1}^{q_i}l_i\left(\Psi_i(\cdot)^Te_k\right)\sum_{j=1}^{d_i}\left(e_je_j^T\right)\left(\Psi_i(*)^Te_k\right)\right) \\
&= l_i\left(\sum_{k=1}^{q_i}l_i\left(\Psi_i(\cdot)^Te_k\right)\left(\Psi_i(*)^Te_k\right)\right) \\
&= \sum_{k=1}^{q_i}l_i\left(\Psi_i(\cdot)^Te_k\right)l_i\left(\Psi_i(*)^Te_k\right) \\
& = \sum_{k=1}^{q_i}(l_i(\Psi(\cdot)^Te_k))^2
\end{align}
Next, let us evaluate
\begin{align}
\left<l^*_i,\mathcal{L}^*_{i,f}\right>_{\mathds{V}_i} &= \mathcal{L}_{i,f}(l^*_i(*)) \\
&= \mathcal{L}_{i,f}\left(\sum_{j=1}^{d_i}l_i(K_{\mathds{V}_i}(\cdot,*)e_j)e_j\right) \\
&=\mathcal{L}_{i,f}\left(\sum_{j=1}^{d_i}l_i(\Psi_i(\cdot)^T\Psi_i(*)e_j)e_j\right) \\
&=\mathcal{L}_{i,f}\left(\sum_{j=1}^{d_i}l_i\left(\Psi_i(\cdot)^T\left(\sum_{k=1}^{q_i}e_ke_k^T\right)\Psi_i(*)e_j\right)e_j\right) \\
&=\mathcal{L}_{i,f}\left(\sum_{j=1}^{d_i}l_i\left(\Psi_i(\cdot)^T\left(\sum_{k=1}^{q_i}e_ke_k^T\right)\Psi_i(*)e_j\right)e_j\right) \\
&=\mathcal{L}_{i,f}\left(\sum_{j=1}^{d_i}\sum_{k=1}^{q_i}l_i\left(\Psi_i(\cdot)^Te_k\left(e_k^T\Psi_i(*)e_j\right)\right)e_j\right) \\
&=\mathcal{L}_{i,f}\left(\sum_{j=1}^{d_i}\sum_{k=1}^{q_i}l_i\left(\Psi_i(\cdot)^Te_k\right)\left(e_k^T\Psi_i(*)e_j\right)e_j\right) \\
&=\mathcal{L}_{i,f}\left(\sum_{j=1}^{d_i}\sum_{k=1}^{q_i}l_i\left(\Psi_i(\cdot)^Te_k\right)e_j\left(e_j^T\Psi_i(*)^Te_k\right)\right) \\
&=\mathcal{L}_{i,f}\left(\sum_{k=1}^{q_i}l_i\left(\Psi_i(\cdot)^Te_k\right)\left(\Psi_i(*)^Te_k\right)\right) \\
&=\sum_{k=1}^{q_i}\mathcal{L}_{i,f}\left(\left(\Psi_i(*)^Te_k\right)\right)l_i\left(\Psi_i(\cdot)^Te_k\right) \\
&=\sum_{k=1}^{q_i}p_i\left(f(*),\Psi_i^T(*)e_k\right)l_i\left(\Psi_i(\cdot)^Te_k\right)\\
\end{align}

Finally, 
\begin{align}
\left<\mathcal{L}^*_{i,f},\mathcal{L}^*_{i,f}\right>_{\mathds{V}_i} &= \mathcal{L}_{i,f}(\mathcal{L}^*_{i,f}(*)) \\
&= \mathcal{L}_{i,f}\left(\sum_{j=1}^{d_i}p_i(f(\cdot),K_{\mathds{V}_i}(\cdot,*)e_j)e_j\right) \\
&= \mathcal{L}_{i,f}\left(\sum_{j=1}^{d_i}p_i(f(\cdot),\Psi_i(\cdot)^T\Psi_i(*)e_j)e_j\right) \\
&= \mathcal{L}_{i,f}\left(\sum_{j=1}^{d_i}p_i\left(f(\cdot),\Psi_i(\cdot)^T\left(\sum_{k=1}^{q_i}e_ke_k^T\right)\Psi_i(*)e_j\right)e_j\right) \\
&= \mathcal{L}_{i,f}\left(\sum_{j=1}^{d_i}\sum_{k=1}^{q_i}p_i\left(f(\cdot),\Psi_i(\cdot)^Te_k\right)\left(e_k^T\Psi_i(*)e_j\right)e_j\right) \\
&= \mathcal{L}_{i,f}\left(\sum_{j=1}^{d_i}\sum_{k=1}^{q_i}p_i\left(f(\cdot),\Psi_i(\cdot)^Te_k\right)e_j\left(e_j^T\Psi_i(*)^Te_k\right)\right) \\
&= \mathcal{L}_{i,f}\left(\sum_{k=1}^{q_i}p_i\left(f(\cdot),\Psi_i(\cdot)^Te_k\right)\left(\Psi_i(*)^Te_k\right)\right) \\
&= \sum_{k=1}^{q_i}\left(p_i\left(f(\cdot),\Psi_i(\cdot)^Te_k\right)\mathcal{L}_{i,f}\left(\Psi_i(*)^Te_k\right)\right) \\
&=\sum_{k=1}^{q_i}\left(p_i\left(f(\cdot),\Psi_i(\cdot)^Te_k\right)\right)^2 
\end{align}
Our loss is then
\begin{multline}
\inf_{f\in \mathds{H}}\left\{\sum_{i=1}^{n}\sum_{k=1}^{q_i}\left(p_i\left(f(*),\Psi_i^T(*)e_k\right)^2 - 2p_i\left(f(*),\Psi_i^T(*)e_k\right)l_i\left(\Psi_i(\cdot)^Te_k\right) \right.\right.\\ \left.\left.+ (l_i(\Psi(\cdot)^Te_k))^2\right)+\lambda(\|f\|^2_{\mathds{H}})\right\}
\end{multline}
which is
\begin{equation}
\inf_{f\in \mathds{H}}\left\{\sum_{i=1}^{n}\sum_{k=1}^{q_i}\left(p_i\left(f(*),\Psi_i^T(*)e_k\right) - l_i\left(\Psi_i(*)^Te_k\right)\right)^2+\lambda(\|f\|^2_{\mathds{H}})\right\}
\end{equation}
\textbf{Part 3:}
Now we use Proposition \ref{thm:Riesz_2} once again, recalling that $p_i$ is continuous with respect to $f$ for any choice of $v$ to rewrite the loss as
\begin{equation}
\inf_{f\in\mathds{H}}\left\{\sum_{i=1}^{n}\sum_{k=1}^{q_i}\left(\left<f,\mathcal{L}^*_{i,k}\right>_{\mathds{H}} - l_i(\Psi_i(*)^Te_k)\right)^2 + \lambda(\|f\|^2_{\mathds{H}})\right\}
\end{equation}
where 
\begin{equation}
\mathcal{L}^*_{i,k}(*) = \sum_{j=1}^{d}p_i(K_{\mathds{H}}(\cdot,*)e_j,\Psi_i(\cdot)^Te_k)e_j
\end{equation}
We may use a representer theorem argument to conclude that 
\begin{equation}
f\in span\left\{\mathcal{L}^*_{i,k}\right\}
\end{equation}
Indeed, letting 
\begin{equation}
f^* = \sum_{i=1}^{n}\sum_{k=1}^{q_i}\alpha_{i,k}\mathcal{L}^{*}_{i,k}
\end{equation}
and 
\begin{equation}
f = f^* + f^{\perp}
\end{equation}
we get:
\begin{align}
&\inf_{f\in\mathds{H}}\left\{\sum_{i=1}^{n}\sum_{k=1}^{q_i}\left(\left<f,\mathcal{L}^*_{i,k}\right>_{\mathds{H}} - l_i(\Psi_i(*)^Te_k)\right)^2 + \lambda(\|f\|^2_{\mathds{H}})\right\}\\
&\Leftrightarrow \inf_{f\in\mathds{H}}\left\{\sum_{i=1}^{n}\sum_{k=1}^{q_i}\left(\left<f^*+f^{\perp},\mathcal{L}^*_{i,k}\right>_{\mathds{H}} - l_i(\Psi_i(*)^Te_k)\right)^2 + \lambda(\|f^*+f^{\perp}\|^2_{\mathds{H}})\right\}\\
& \Leftrightarrow \inf_{f\in\mathds{H}}\left\{\sum_{i=1}^{n}\sum_{k=1}^{q_i}\left(\left<f^*,\mathcal{L}^*_{i,k}\right>_{\mathds{H}} - l_i(\Psi_i(*)^Te_k)\right)^2 + \lambda(\|f^*+f^{\perp}\|^2_{\mathds{H}})\right\}\\
& \geq  \inf_{f^*\in\mathds{H}}\left\{\sum_{i=1}^{n}\sum_{k=1}^{q_i}\left(\left<f^*,\mathcal{L}^*_{i,k}\right>_{\mathds{H}} - l_i(\Psi_i(*)^Te_k)\right)^2 + \lambda(\|f^*\|^2_{\mathds{H}})\right\}\\
\end{align}
If $\lambda$ is a strictly increasing function we get a strict inequality and uniqueness of the solution. Thus we get the finite dimensional optimization problem
\begin{small}
\begin{multline}
\inf_{\alpha_{l,m}}\left\{\sum_{i=1}^{n}\sum_{k=1}^{q_i}\left(\sum_{l=1}^{n}\sum_{m=1}^{q_l}\alpha_{l,m}\left<\mathcal{L}^*_{l,m},\mathcal{L}^{*}_{i,k}\right>_{\mathds{H}}-l_i(\Psi_i(\cdot)^Te_k)\right)^2\right.\\\left. + \lambda\left(\sum_{l=1}^{n}\sum_{m=1}^{q_l}\sum_{i=1}^{n}\sum_{k=1}^{q_i}\alpha_{l,m}\alpha_{i,k}\left<\mathcal{L}^*_{l,m},\mathcal{L}^*_{i,k}\right>_{\mathds{H}}\right)\right\}
\end{multline}
\end{small}
Assuming $\lambda(\|f\|^2_{\mathds{H}}) = \lambda \|f\|_{\mathds{H}}^2$ and $N = \sum_{i=1}^{n}q_i$. 
Letting $M \in \mathds{R}^{N \times N}$ with
\begin{align}
M_{l,m,i,k} &= \left<\mathcal{L}^*_{l,m},\mathcal{L}^*_{i,k}\right>_{\mathds{H}}\\
\left<\mathcal{L}^*_{l,m},\mathcal{L}^*_{i,k}\right>_{\mathds{H}} &= \mathcal{L}_{i,k}(\mathcal{L}^*_{l,m}(*)) \\
&= \mathcal{L}_{i,k}\left(\sum_{j=1}^{d}p_l\left(K_{\mathds{H}}(\cdot,*)e_j,\Psi_{l}(\cdot)^Te_m\right)e_j\right) \\
&=\mathcal{L}_{i,k}\left(\sum_{j=1}^{d}p_l\left(K_{\mathds{H}}(\cdot,*)e_j,\Psi_{l}(\cdot)^Te_m\right)e_j\right) \\
&= p_i\left(\sum_{j=1}^{d}p_l\left(K_{\mathds{H}}(\cdot,*)e_j,\Psi_{l}(\cdot)^Te_m\right)e_j,\Psi_{i}(*)^Te_k\right) \\
\end{align}
Notice, unless we know the bilinear form is defined for functions of the form
\begin{equation}
    p_l\left(K_{\mathds{H}}(\cdot,*)e_j,\Psi_l(\cdot)^Te_m\right)
\end{equation}
we can't pull the sum outside the bilinear form. In practice, after the bilinear form is defined we can move the sum outside. In all other cases, we are careful to only move the sum outside if the term on the inside is an element of the RKHS over which the linear functionals or bilinear functionals are defined. Letting $y \in \mathds{R}^{N}$ with
\begin{equation}
y_{i,k} = l_i(\Psi_i(\cdot)^Te_k)
\end{equation}
we get $\alpha \in \mathds{R}^{N}$ by solving the linear system
\begin{equation}
\left(M+\lambda I\right)\alpha = y
\end{equation}
\end{Proof}

\section{Ridge regression example} \label{app: ridge}
Assuming we wish to minimize:
\begin{equation}
\min_{f\in\mathds{H}}\left\{\sum_{i=1
}^n\|f(x_i) - y_i\|^2 + \|f\|_{\mathds{H}}^2\right\}
\end{equation}
where $\mathds{H}$ is a vector valued RKHS with matrix valued kernel $K_{\mathds{H}}(\cdot,*) \in \mathds{R}^{p\times p}$. The standard ridge regression solution is
\begin{equation}
f(x) = \left[K_{\mathds{H}}(x,x_1),...,K_{\mathds{H}}(x,x_n)\right]\alpha
\end{equation}
with 
\begin{equation}
\alpha = \left(M + \lambda I\right)^{-1}Y
\end{equation}
with 
\begin{equation}
M = M^T = \left[\begin{array}{ccc}
K_{\mathds{H}}(x_1,x_1)&...&K_{\mathds{H}}(x_1,x_n)\\
\vdots&\ddots&\vdots\\
K_{\mathds{H}}(x_n,x_1)&...&K_{\mathds{H}}(x_n,x_n)\end{array}\right]
\end{equation}
and 
\begin{equation}
Y = \left[\begin{array}{c}y_1\\\vdots \\ y_n\end{array}\right]
\end{equation}
Treating the problem as a variational problem we minimize the loss:
\begin{equation}
\inf_{f\in\mathds{H}}\left\{\sum_{i=1}^{n}\sum_{j=1}^{p}\sup_{v\in\mathds{V}_{i,j},\|v\|=1}\left\{(p_{i,j}(f,v) - l_{i,j}(v))^2\right\}+\lambda\|f\|^2\right\}
\end{equation}
Where $\mathds{V}_{i,j}$ is a vector valued RKHS with the (somewhat trivial) constant kernel $K(\cdot,*) = e_je_j^T = \Psi_{i,j}(\cdot)^T\Psi_{i,j}(*)$ with $\Psi_{i,j}(\cdot) = e_j^T$. 
\begin{equation}
    p_{i,j}(f,v) = f(x_i)^Tv(x_i) = f(x_i)^Te_j
\end{equation}
and
\begin{equation}
l_{i,j}(v) = y_i^Tv(x_i) = y_i^Te_j
\end{equation}
Theorem\ref{thm:BigThm_sec} allows us to conclude 
\begin{equation}
 f(x) = \sum_{i=1}^{n}\sum_{j=1}^{p}\sum_{k=1}^{1}\alpha_{i,j,k}\mathcal{L}^*_{i,j,k}(x)
\end{equation}
with $k$ the dimension of the feature space (1 in our case) and
\begin{equation}
\mathcal{L}^*_{i,j,k}(*) = \sum_{l=1}^{p}P_{i,j}\left(K_{\mathds{H}}(\cdot,*)e_l,\Psi_{i,j}(\cdot)^Te_k\right)e_l
\end{equation}
Here $\Psi_{i,j}(\cdot)^T = e_j$ and $e_k=1$ so we simplify to
\begin{align}
\mathcal{L}^*_{i,j,k}(*) &= \sum_{l=1}^{p}P_{i,j}(K_{\mathds{H}}(\cdot,*)e_l,e_j)e_l \\
&= \sum_{l=1}^{p}\left[\left(K_{\mathds{H}}(x_i,*)e_l\right)^Te_j\right]e_l\\
&= \sum_{l=1}^{p}\left[e_l^TK_{\mathds{H}}(*,x_i)e_j\right]e_l \\
&= \sum_{l=1}^{p}\left[e_le_l^TK_{\mathds{H}}(*,x_i)e_j\right] \\
&= \left(\sum_{l=1}^{p}e_le_l^T\right)K_{\mathds{H}}(*,x_i)e_j \\
&= K(*,x_i)e_j
\end{align}
Thus
\begin{equation}
f(*) = \sum_{i=1}^{n}\sum_{j=1}^{p}\alpha_{i,j}K_{\mathds{H}}(*,x_i)e_j = \sum_{i=1}^{n} K_{\mathds{H}}(*,x_i)\alpha_i
\end{equation}
This proves $f$ takes the form as in Regression, all that is left to prove is the $\alpha$ we learn is the same as that learned by Regression. For this, recall from theorem \ref{thm:BigThm_sec} that the $\alpha$ we desire to learn satisfies
\begin{equation}
(M+\lambda I)\alpha = Y
\end{equation}
Where
\begin{equation}
M = \left[\begin{array}{ccccccc}
\left<\mathcal{L}^*_{1,1},\mathcal{L}^*_{1,1}\right>&\left<\mathcal{L}^*_{1,2},\mathcal{L}^*_{1,1}\right> &... &\left<\mathcal{L}^*_{1,p},\mathcal{L}^*_{1,1}\right>&\left<\mathcal{L}^*_{2,1},\mathcal{L}^*_{1,1}\right>&...&\left<\mathcal{L}^*_{n,p},\mathcal{L}^*_{1,1}\right> \\
\left<\mathcal{L}^*_{1,1},\mathcal{L}^*_{1,2}\right>&\left<\mathcal{L}^*_{1,2},\mathcal{L}^*_{1,2}\right>&...&\left<\mathcal{L}^*_{1,p},\mathcal{L}^*_{1,2}\right>&\left<\mathcal{L}^*_{2,1},\mathcal{L}^*_{1,2}\right>&...&\left<\mathcal{L}^*_{n,p},\mathcal{L}^*_{1,2}\right> \\
\vdots & \vdots & \ddots & \vdots & \vdots & \ddots & \vdots \\
\left<\mathcal{L}^*_{1,1},\mathcal{L}^*_{1,p}\right>&\left<\mathcal{L}^*_{1,2},\mathcal{L}^*_{1,p}\right>&...&\left<\mathcal{L}^*_{1,p},\mathcal{L}^*_{1,p}\right>&\left<\mathcal{L}^*_{2,1},\mathcal{L}^*_{1,p}\right>&...&\left<\mathcal{L}^*_{n,p},\mathcal{L}^*_{1,p}\right> \\
\left<\mathcal{L}^*_{1,1},\mathcal{L}^*_{2,1}\right>&\left<\mathcal{L}^*_{1,2},\mathcal{L}^*_{2,1}\right>&...&\left<\mathcal{L}^*_{1,p},\mathcal{L}^*_{2,1}\right>&\left<\mathcal{L}^*_{2,1},\mathcal{L}^*_{2,1}\right>&...&\left<\mathcal{L}^*_{n,p},\mathcal{L}^*_{2,1}\right> \\
\vdots & \vdots & \ddots & \vdots & \vdots & \ddots & \vdots \\
\left<\mathcal{L}^*_{1,1},\mathcal{L}^*_{n,p}\right>&\left<\mathcal{L}^*_{1,2},\mathcal{L}^*_{n,p}\right>&...&\left<\mathcal{L}^*_{1,p},\mathcal{L}^*_{n,p}\right>&\left<\mathcal{L}^*_{2,1},\mathcal{L}^*_{n,p}\right>&...&\left<\mathcal{L}^*_{n,p},\mathcal{L}^*_{n,p}\right> \\
\end{array}
\right]
\end{equation}
and 
\begin{equation}
Y = \left[\begin{array}{c}
l_{1,1}(\Psi_{1,1}(\cdot)^T)\\
l_{1,2}(\Psi_{1,2}(\cdot)^T) \\
\vdots \\
l_{n,d}(\Psi_{n,p}(\cdot)^T)
\end{array}\right]
\end{equation}
Note, since the feature function is one dimensional $e_k = 1 \in \mathds{R}$ simplifying matters.
Starting with $Y$ we observe:
\begin{equation}
l_{i,j}(\Psi_{i,j}(\cdot)^T) = l_{i,j}(e_j) = y_i^Te_j
\end{equation}
Thus
\begin{equation}
Y = \left[\begin{array}{c}
y_1 \\
\vdots \\
y_n
\end{array}\right]
\end{equation}
as desired. Finally,
\begin{align}
\left<\mathcal{L}^*_{i,j},\mathcal{L}^*_{k,l}\right> &= p_{i,j}\left\{\sum_{m=1}^{p}p_{k,l}\left(K_{\mathds{H}}(\cdot,*)e_m,\Psi_{k,l}(\cdot)^T\cdot 1\right)e_m,\Psi_{i,j}(\cdot)^T\cdot 1\right\}\\
&= p_{i,j}\left\{\sum_{m=1}^{p}p_{k,l}\left(K_{\mathds{H}}(\cdot,*)e_m,e_l\right)e_m,e_j\right\}\\
&= p_{i,j}\left\{\sum_{m=1}^{p}\left[\left(K_{\mathds{H}}(x_k,*)e_m\right)^Te_l\right]e_m,e_j\right\}\\
&= p_{i,j}\left\{\sum_{m=1}^{p}\left[e_m^TK_{\mathds{H}}(*,x_k)e_l\right]e_m,e_j\right\}\\
&= p_{i,j}\left\{K_{\mathds{H}}(*,x_k)e_l,e_j\right\}\\
&= e_l^TK_{\mathds{H}}(x_k,x_i)e_j
\end{align}
Thus, $M$ is also of the desired form. We note that once again we use the fact that the feature function is one dimensional.
\section{Liouville operator occupation kernel generalization}
\begin{multline}
\label{eq:luivOCK}
\inf_{u\in U}\left\|\int_{0}^T \nabla_{2}\left(k(\cdot,\gamma(t))\right)^Tf(\gamma(t))dt - \int_{0}^T\nabla_{2}\left(k(\cdot,\gamma(t))\right)^Tu(\gamma(t))dt\right\|^2_{\mathds{V}} \Leftrightarrow \\
\inf_{u\in U}\left\{\sup_{v\in \mathds{V}, \|v\|=1}\left\{\left(\int_{0}^T \nabla\left(v(\gamma(t))\right)^Tf(\gamma(t))dt - \int_{0}^T\nabla\left(v(\gamma(t))\right)^Tu(\gamma(t))dt\right)^2\right\}\right\}
\end{multline}
Since 
\begin{equation}
\int_0^T \nabla(v(\gamma(t)))^Tf(\gamma(t))dt
\end{equation}
is a bounded bilinear functional with respect to $v$, and
\begin{equation}
\int_{0}^T\nabla (v(\gamma(t)))^Tu(\gamma(t))dt
\end{equation}
is a bounded bilinear form with respect to $v$ and $u$, Reisz representer theorem guarantees occupation kernel representers of these linear functionals in $\mathds{V}$. These may be derived as in theorem \ref{thm:Riesz_2}. Consider:
\begin{equation}
l(v) \equiv \int_0^T \nabla(v(\gamma(t)))^Tf(\gamma(t))dt
\end{equation}
We seek
\begin{equation}
l^*(x)
\end{equation}
for which
\begin{equation}
\left<v,l^*\right> = l(v)
\end{equation}
for every $v$. Observe
\begin{align}
l^*(x) &= \left<l^*, k(\cdot,x)\right> \\
 &= \left<k(\cdot,x),l^*\right> \\
 &= l(k(\cdot,x)) \\
 &= \int_{0}^T \nabla_2\left(k(x,\gamma(t))\right)^Tf(\gamma(t))dt 
\end{align}
Similarly, letting
\begin{equation}
    \mathcal{L}(v) = \int_{0}^T\nabla(v(\gamma(t)))u(\gamma(t))dt
\end{equation}
we get 
\begin{equation}
\mathcal{L}^*(x) = \int_{0}^T \nabla_2\left(k(x,\gamma(t))\right)^Tu(\gamma(t))dt 
\end{equation}
and we rewrite the supremum as
\begin{align}
&\sup_{v\in \mathds{V}, \|v\|=1}\left\{\left(\int_{0}^T \nabla\left(v(\gamma(t))\right)^Tf(\gamma(t))dt - \int_{0}^T\nabla\left(v(\gamma(t))\right)^Tu(\gamma(t))dt\right)^2\right\} \\
&= \sup_{v\in \mathds{V}, \|v\|=1}\left\{\left(\left<v, \int_{0}^T \nabla_2\left(k(\cdot,\gamma(t))\right)^Tf(\gamma(t))dt-\int_{0}^T \nabla_2\left(k(\cdot,\gamma(t))\right)^Tu(\gamma(t))dt\right>\right)^2\right\} \\
&= \left\|\int_{0}^T \nabla_2\left(k(\cdot,\gamma(t))\right)^Tf(\gamma(t))dt - \int_{0}^T \nabla_2\left(k(\cdot,\gamma(t))\right)^Tu(\gamma(t))dt\right\|^2_{\mathds{V}}
\end{align}
\section{Derivation of ROCK for the PDE model}\label{PDE_supp}
Letting $u_i = u(t,x_i)$, The cost function given in~\eqref{pde_cost_1} is an application of:
\begin{samepage}
\begin{small}
    \begin{multline}\label{ROCK-PDE}
    \min_{f \in \mathds{H}}\left\{\sum_{i=1}^{N} \sum_{j=1}^{M-1} \left\{\sup_{v\in \mathds{V}, \|v\|=1}\left(\int_{t_o}^{t_f} f(u_i, \partial_x u_i, \partial_{xx} u_i, \dots) v(t,x_i)dt\right.\right.\right.\\\left.\left.\left. - \int_{t_o}^{t_f} \partial_t u(t,x_i)  v(t,x_i)dt \right)^2\right\} + \lambda\|f\|^2\right\}
\end{multline}
\end{small}
    
\end{samepage}
where $v(t,x_i)$ are test functions. Choosing constant test functions $v$ on the intervals $[t_j,t_{j+1}]$, and observing that these spaces are one dimensional (so the sup is over a trivial set of two functions). We get the transformed problem:
\begin{equation}
\min_{f\in\mathds{H}}\left\{\sum_{i=1}^{N}\sum_{j=1}^{M-1}\left\{\left(\int_{t_j}^{t_{j+1}}f(u_i,\partial_xu_i,...)dt - \int_{t_j}^{t_{j+1}}\partial_t u_idt\right)^2\right\}+\lambda\|f\|^2\right\}
\end{equation}
The fundamental theorem of calculus allows us to arrive at the loss used in our implementation:
\begin{equation}
\min_{f\in\mathds{H}}\left\{\sum_{i=1}^{N}\sum_{j=1}^{M-1}\left\{\left(\int_{t_j}^{t_{j+1}}f(u_i,\partial_xu_i,...)dt - u(t_{j+1},x_i) + u(t_j,x_i)\right)^2\right\}+\lambda\|f\|^2\right\}\end{equation}

\section{Legendre polynomials}
\label{app:legendre}
Defining $T_0(x) = 1$ and  $T_1(x)= x$, the Legendre polynomials can be determined by the three term recurrence relation 
\begin{equation}
    (n+1)T_{n+1}(x) = (2n+1)xT_{n}(x) - nT_{n-1}(x).
\end{equation}
Below we provide the first few Legendre polynomials on the right along with their $L^2([-1,1])$ norm on the left. In practice, we shift and scale these functions to define subspaces of $L^2([a,b])$ for time intervals $[a,b]$.
\begin{align}
\sqrt{2} & &T_0(x)= &1& \\
\sqrt{\frac{2}{3}} & & T_1(x)= &x& \\
\sqrt{\frac{2}{5}} & &T_2(x)= &\frac{1}{2}\left(3x^2-1\right)& \\
\sqrt{\frac{2}{7}} & &T_3(x)= &\frac{1}{2}\left(5x^3-3x\right)& \\
\sqrt{\frac{2}{9}} & &T_4(x)= & \frac{1}{8}\left(35x^4 - 30x^2 + 3\right)& \\
\sqrt{\frac{2}{11}} & &T_5(x)= & \frac{1}{8}\left(63x^5 - 70x^3 + 15x\right) 
%\vdots &  & \vdots 
\end{align}
%Where the term on the left denotes the $L^2([-1,1])$ norm of the Legendre polynomial on the right. We shift and scale these functions to define subspaces of $L^2([a,b])$ for time intervals $[a,b]$.

\section{ROCK implementation}\label{app: implementation}
\subsection{Vectorizations and optimizations}\label{app:Vectorizations}
Vectorization (and the use of properties of Kronecker products) is the heart of what allows the ROCK algorithm to be so remarkably concise and practically efficient to run. We expect similar optimizations may be applied to many other losses derived from our framework. \\
Recall for ROCK we need to solve the linear system
\begin{equation}
(M+\lambda I)\alpha = y
\end{equation}
where
\begin{equation}
y \in \mathds{R}^{npd}
\end{equation}
with
\begin{equation}
y_{i} = \int_{a^i}^{b^i}\Psi_i(t)\dot{x}^i(t)dt = \int_{a^i}^{b^i}\left(\psi_i(t)\otimes I\right)\dot{x}^i(t)dt
\end{equation}
On the other hand, the $i,i'$th block of the matrix $M$ is given by
\begin{align}
&\int^{b^i}_{a^i}\int_{a^{i'}}^{b^{i'}}\Psi_{i'}(\tau)K_{\mathds{H}}(x^{i'}(\tau),x^{i}(t))\Psi_{i}(t)^Td\tau dt \\
&= \int^{b^i}_{a^i}\int_{a^{i'}}^{b^{i'}}\left(\psi_{i'}(\tau)\otimes I\right)\left(k(x^{i'}(\tau),x^{i}(t))I\right)\left(\psi_{i}(t)\otimes I\right)^Td\tau \\
&=  \int^{b^i}_{a^i}\int_{a^{i'}}^{b^{i'}}k(x^{i'}(\tau),x^{i}(t))\left(\psi_{i'}(\tau)\otimes I\right)\left(\psi_{i}(t)\otimes I\right)^Tdtd\tau  \\
&= \int^{b^i}_{a^i}\int_{a^{i'}}^{b^{i'}}k(x^{i'}(\tau),x^{i}(t))\left(\psi_{i'}(\tau)\psi_{i}(t)^T\otimes  I\right)dtd\tau \\
&= \left(\int^{b^i}_{a^i}\int_{a^{i'}}^{b^{i'}}\left(k(x^{i'}(\tau),x^{i}(t))\psi_{i'}(\tau)\psi_{i}(t)^T\right)dtd\tau\right) \otimes I\\
&\approx \left(\sum_{k=1}^{n}\sum_{j=1}^{m}q_kq_j\left(k(x^{i'}(\tau_k),x^{i}(t_j))\psi_{i'}(\tau_k)\psi_{i}(t_j)^T\right)\right) \otimes I
\end{align}
where $q_k$ and $q_j$ are weights derived from the choice of numerical quadrature. If we let $K^{i',i}$ be the gram matrix $K^{i',i}_{j:k} = k(x^{i'}(\tau_k),x^i(t_j))$, and $q\Psi_t$ and $q\Psi_{\tau}$ be matrices associated with the feature functions and quadratures $[q\Psi_{\tau}]_{k} = q_k \psi_{i'}(\tau_k)\in \mathds{R}^q$ then we can estimate all the terms of the integral by evaluating the matrix multiplications
\begin{equation}
G_{i',i} = (q\Psi_{\tau}) K^{i',i} (q\Psi_{t})^T
\end{equation}
In addition, a linear system of the form
\begin{equation}
\left(M\otimes I\right)\alpha = y
\end{equation}
such as our linear system, may be rewritten using properties of Kronecker products as
\begin{equation}
M A = Y
\end{equation}
where $A$ and $Y$ are matrix valued and $\alpha$ and $y$ are flattened versions of $A$ and $Y$. This allows us to solve a much smaller linear system. $$(G+\lambda I)A = Y$$

\subsection{Practical implementation}
We employ some vectorization techniques to arrive at a remarkably concise and yet very general implementation of the ROCK method as follows: We note that theorem \ref{thm:BigThm_sec} requires us to evaluate 
\begin{equation}
Y, \text{ and } \ M
\end{equation}
from which we calculate 
\begin{equation}
\alpha
\end{equation}
At which point we may evaluate $f$ at new datapoints. For the following we assume the $n$th trajectory has $m_i$ samples for us to use in our integral estimate. Recall:
\begin{equation}
Y \in \mathds{R}^{pdn}
\end{equation}
where for $i\in [1,...,n]$ and $j\in [1,...,p]$ 
\begin{equation}
y_{i,j} = x_i(b_i)[\psi_i(b_i)]_j - x_i(a_i)[\psi_i(a_i)]_j - \int_{a_i}^{b_i}x_i(t)[\dot{\psi}_i(t)]_jdt \in \mathds{R}^d
\end{equation}
here, $[\psi_i(b_i)]_j$ and $[\dot{\psi}_i(t)]_j$ are scalar valued, and $x_i(t)\in \mathds{R}^d$.
We calculate $\Psi \in \mathds{R}^{np \times (\sum_{i=1}^{n}m_i)}$ a block diagonal matrix of test function evaluations with the blocks:
\begin{equation}
\psi_{i,i} = \left[\begin{array}{cccc}
\psi_i(t^i_1), &
\psi_i(t^i_2), &
..., &
\psi_i(t^i_{m_i})\end{array}\right] \in \mathds{R}^{p\times m_i}
\end{equation}
We calculate $\Psi$ in a for loop through each of the n trajectories. This is the only part of the algorithm that is not fully vectorized, however, we choose to implement this in this way to allow for the freedom of using different test functions for each trajectory. We find this part of the algorithm to be insignificant in terms of runtime and memory costs. Each block is calculated in a vectorized way with the following python code. (We implement Legendre polynomials over arbitrary intervals as our test functions as described Section \ref{sec:Experiments}).
\begin{python}
# We will be using Scaled and Shifted Legendre Polynomials
# ts - The array of times to evaluate the Legendre polynomials at
# n  - the order of the polynomial (integer with 0 <= n).
# a  - The left end of the interval to shift the polynomials to
# b  - the right end of the interval to shift the polynomials to. (b > a)
# Computes the Legendre Polynomials based on the recursion:
# P_{n+1}(x) = [(2n+1)xP_{n}(x) - n*P_{n-1}(x)]/(n+1)
def phis(ts,n,a,b):
  # Apply the transformation of ts to tsp (on the interval [-1,1])
  if n == 0:
    Ts = np.ones((len(ts),1))
    return Ts
  tsp = (2/(b-a))*ts - (b+a)/(b-a)
  Ts = np.ones((len(ts),n+1))
  Ts[:,1] = tsp
  for i in range(2,n+1):
    Ts[:,i] = ((2*(i-1)+1)*tsp*Ts[:,i-1] - (i-1)*Ts[:,i-2])/(i)
  # The normalization term for the scaled Legendre Polynomial is
  # sqrt((2*n+1)/2)*sqrt(2/(b-a))
  Ts = Ts*np.sqrt((np.array(range(n+1))*2+1)/2)[None,:]*np.sqrt(2/(b-a))
  return Ts
\end{python}
In addition, we compute 
$\dot{\Psi}_{i,i}\in \mathds{R}^{p\times m_i}$, a matrix of derivatives of the test functions. Code for calculating both $\Psi_{i,i}$ and $\dot{\Psi}_{i,i}$ in one function is presented below:
\begin{python}
# We will be using Scaled and Shifted Legendre Polynomials
# We need the derivatives as well.
# ts - The array of times to evaluate the Legendre polynomials at
# n  - the order of the polynomial (integer with 0 <= n).
# a  - The left end of the interval to shift the polynomials to
# b  - the right end of the interval to shift the polynomials to. (b > a)
# Computes the derivative of the Legendre Polynomials based on the recursion:
# P_{n+1}(x) = [(2n+1)xP_{n}(x) - n*P_{n-1}(x)]/(n+1), which implies:
# P'_{n+1}(x) = [(2n+1)P_{n}(x) + (2n+1)xP'_{n}(x)-n*P'_{n-1}(x)]/(n+1)
def phisp(ts,n,a,b):
  # Apply the transformation of ts to tsp (on the interval [-1,1])
  tsp = (2/(b-a))*ts - (b+a)/(b-a)
  if n == 0:
    return np.ones((len(ts),1)), np.zeros((len(ts),1))
  Ts = np.ones((len(ts),n+1))
  Tsp = np.zeros((len(ts),n+1))
  Ts[:,1] = tsp
  Tsp[:,1] = 1
  for i in range(2,n+1):
    Ts[:,i]=((2*(i-1)+1)*tsp*Ts[:,i-1] - (i-1)*Ts[:,i-2])/(i)
    Tsp[:,i]=((2*(i-1)+1)*Ts[:,i-1]+(2*(i-1)+1)*tsp*Tsp[:,i-1]-(i-1)*Tsp[:,i-2])/(i)
  # The normalization term for the scaled Legendre Polynomial is
  # sqrt((2*n+1)/2)*sqrt(2/(b-a))
  Ts = Ts*np.sqrt((np.array(range(n+1))*2+1)/2)[None,:]*np.sqrt(2/(b-a))
  Tsp = Tsp*np.sqrt((np.array(range(n+1))*2+1)/2)[None,:]*np.sqrt(2/(b-a))*(2/(b-a))
  return Ts,Tsp
\end{python}
We construct $QPhi\in \mathds{R}^{pn\times (\sum_{i=1}^{n}m_i)}$ and $QPhiD \in \mathds{R}^{pn\times(\sum_{i=1}^{n}m_i)}$. These are both block diagonal matrices with the blocks:
\begin{equation}
QPhi_{i,i} = [q^i_1\Psi_i(t^i_1),q^i_2\Psi_i(t^i_2),...,q^i_{m_i}\Psi_i(t^i_{m_i})]\in \mathds{R}^{p\times m_i} 
\end{equation}
where $q^i_j$ is the quadrature weight used to estimate the integral using the time points $t^i_1,...,t^i_{m_i}$. For instance, assuming $t^i_j = jh$, and using a composite trapezoid rule, $q^i_1 = h/2$, $q^i_2 = ... = q^i_{m_i-1} = h, q^i_{m_i} = h/2$.
\begin{multline}
    QPhiD_{i,i} = [-\Psi_i(t^i_1)-q^i_1\dot{\Psi}_i(t^i_1),-q^i_2\dot{\Psi}_i(t^i_2) \\ ,...,-q^i_{m_i-1}\dot{\Psi}_i(t^i_{m_i-1}),\Psi_i(t^i_{m_i}) - q^i_{m_i}\dot{\Psi}_i(t^i_{m_i})]\in \mathds{R}^{p\times m_i}
\end{multline}
Here we notice that in addition to the derivative terms, $QPhiD$ also involves the boundary terms on the first and last columns of each block. These block diagonal matrices are constructed in the following code (which contains the only for loop in the algorithm implementation).
\begin{python}
# Requires a list of ts a list of qs, and outputs a block diagonal matrix
# corresponding to the QPhi block diagonal matrix and QPhiD block diagonal matrix.
# Probably the least efficient part of the code. (loops through the trajectories)
def Calculate_QPhi(qs,ts,n):
  psArr = []
  pspArr = []
  for i in range(len(ts)):
    ps, psp = phisp(ts[i],n,ts[i][0],ts[i][-1])
    psp = -(psp[:]*qs[i][:,None])
    # Add boundary terms to psp
    psp[0,:] = psp[0,:]-ps[0,:]
    psp[-1,:] = psp[-1,:]+ps[-1,:]
    psp = psp.T
    ps = (ps*qs[i][:,None]).T
    psArr.append(ps)
    pspArr.append(psp)
  return sp.block_diag(psArr), sp.block_diag(pspArr)
\end{python}
Assuming $dataX \in \mathds{R}^{d\times (\sum_{i=1}^{n}m_i)}$ is the dataset with trajectory $i$ given by:
\begin{equation}
[x_i(t^i_1),...,x_i(t^i_{m_i})] = dataX\left[:,\left(\sum_{k=1}^{i-1}m_k:\sum_{k=1}^{i}m_k\right)\right]
\end{equation}
Then we may calculate $Y$ with:
$Y = dataX \cdot QPhiD^T \in \mathds{R}^{d\times np}$ where $\cdot$ is matrix matrix multiplication.
\begin{python}
# Calculate Y, requires QPhi, QPhiD (derivatives of QPhi), data, and qs (to recover
# Phi from QPhi)
def Calculate_y(QPhiD,qs,dataX):
  ys = dataX@(QPhiD.T)
  return ys
\end{python}
We use a gaussian kernel for our implicit scalar valued kernel. We calculate the gram matrix between two (possibly different datasets $X\in \mathds{R}^{d\times M}, \ and \ Y\in\mathds{R}^{d\times N}$) with
\begin{python}
def gaussianK(X,Y,sigma=kernSig):
  X2 = np.sum(X*X,axis=0)
  Y2 = np.sum(Y*Y,axis=0)
  XY = np.dot(X.T,Y)
  # -X2 - Y2 + 2XY = ||X-Y||^2
  # f(||X-Y||), f(np.sqrt(-X2 - Y2 + 2XY))
  return np.exp((-X2[:,None]-Y2[None,:]+2*XY)/(2*sigma**2))
\end{python}
M is then calculated using:
\begin{equation}
    M = QPhi\cdot (G\cdot QPhi^T)
\end{equation}
Where $G$ is the gram matrix between two datasets $dataX$ and $dataY$. For M in the algorithm, $dataX = dataY$. It is interesting to note that in our algorithms we make heavy use of a connection between integration and matrix matrix multiplication. This allows us to implement the integrations in a provably optimal way in terms of runtime.
\begin{python}
# QPhi is the block diagonal matrix computed by the Calculate_QPhi function
def Calculate_M(QPhi,dataX,dataY,kernel=lambda X, Y : gaussianK(X,Y,sigma=kernSig)):
  G = kernel(dataX,dataY)
  return QPhi@(G@(QPhi.T))
\end{python}
Using properties of Kronecker products, we can solve the linear system in parallel by solving
\begin{equation}
(M + \lambda I)\alpha = Y^T
\end{equation}
where $\alpha \in \mathds{R}^{np\times d}$
We get this with the code:

\begin{python}
# dataset   - (dataX in R^{d x sum_{i} m_i})
# ts        - (list of time vectors for dataset (could be different lengths))
# qs        - (list of time vectors of quadratures (could be different lengths))
# n         - (number of features in the test space)
# lambdaReg - (Regularization term for the alpha)
# sigma     - (The scale parameter for the kernels)
# gaussianK - (A pointer to the implicit kernels)
def learn_alpha(dataset,ts,qs,n,lambdaReg=.0000001,sigma=kernSig,kern = gaussianK):
  Qphi, QphiD = Calculate_QPhi(qs,ts,n)
  M = Calculate_M(Qphi,dataset,dataset,kernel= lambda X, Y : kern(X,Y, sigma=sigma))
  ys = Calculate_y(QphiD,qs,dataset)
  alphas = np.linalg.solve(M+lambdaReg*np.eye(M.shape[0]),ys.T)
  return alphas, Qphi
\end{python}
We evaluate $f$ at new datapoints with the almost the same code as we use to calculate $M$:
\begin{python}
# Calculate f, requires Qhi, alpha, and the data to evaluate f at (dataX) as
# well as the data of the training set (dataY). (alpha should be matrix valued)
def Calculate_f(QPhi,dataX,dataY,alpha,kernel = gaussianK, sigma= kernSig):
  G = kernel(dataX,dataY,sigma)
  return alpha.T@((G@(QPhi.T)).T)
\end{python}
%%%%%%%%%%%%%%%%%%%%%%%%%%%%%%%%%%%%%%%%%%%%%%%%%%%%%%%%%%%%

\section{ROCK performance and runtime considerations}
\label{app:Computational}

\label{app:comp results}

\begin{table}[h!]
\begin{center}
\begin{footnotesize}
$$\begin{array}{c|c|ccccccccc}
&Length & 2 & 5 & 10 & 15 & 25 & 50 & 75 & 300 & 600 \\\hline
&   1   & 9.8M & 3.9M & 2.0M & 1.3M & .79M & .39M & .26M & 66K & 33K\\
&   2  & 20M & 7.8M & 4.0M & 2.6M & 1.6M & .78M & .52M & .13M & 66K\\
&      3  & 29M& 12M& 6.0M & 3.9M & 2.4M & 1.2M & .78M & .20M & 99K\\
features & 4  & 39M & 16M & 8.0M & 5.2M & 3.2M & 1.6M & 1.04M & .26M & .13M\\
& 5  & 49M & 20M & 10M & 6.5M & 4.0M & 2.0M & 1.3M & .33M & .17M\\
& 6  & 59M & 23M & 12M & 7.8M & 4.7M & 2.3M & 1.6M & .40M & .20M\\
& 7  & 69M & 27M & 14M & 9.1M & 5.5M & 2.7M & 1.8M & .46M & .23M\\\hline
\end{array}$$
\end{footnotesize}
\end{center}
\caption{The number of parameters (K denotes thousands M millions) in the ROCK model for each choice of trajectory length in the training set and test feature number for the Kolmogorov dataset. The number of trajectories in the training set is given as the ratio of 1200 and the length of the trajectories. Notice all other datasets will have similar parameter counts (after scaling by the number of dimensions)}
 %The number of parameters (K denotes thousands M millions) in the ROCK model for each choice of trajectory length in the training set and test feature number for the Kolmogorov dataset. The number of trajectories in the training set is given as the ratio of 1200 and the length of the trajectories. Notice all other datasets will have similar parameter counts (after scaling by the number of dimensions)
\label{table:numparams}
\end{table}
\begin{table}[h!]
\begin{center}
\begin{footnotesize}
$$\begin{array}{c|c|ccccccccc}
&Train \ length & 2 & 5 & 10 & 15 & 25 & 50 & 75 & 300 & 600 \\\hline
&1    & 56 & 57 & 57 & 58 & 61 & 56 & 58 & 72 & 73\\
&2   & 58 & 57 & 57 & 57 & 58 & 59 & 54 & 71 & 72\\
&3  & 71  & 65 & 67 & 68 & 56 & 58 & 60 & 70 & 72\\
features&4  & 71 & 71 & 58 & 59 & 63 & 55 & 59 & 59 & 70\\
&5  & 71 & 71 & 55 & 64 & 67 & 55 & 57 & 48 & 69\\
&6  & 71 & 71 & 71 & 71 & 55 & 53 & 54 & 49 & 70\\
&7  & 71 & 71 & 67 & 58 & 58 & 54 & 60 & 58 & 69\\\hline
\end{array}$$
\end{footnotesize}
\end{center}
\caption{The validation performance (E1) of the ROCK model for each choice of trajectory length in the training set and test feature number for the Kolmogorov dataset. On the top left is the validation performance for ROCK with hyper-parameters consistent with the original MOCK implementation (1 feature and 2 sample training trajectories). We see for most choices of features and trajectory lengths we can outperform on the validation set with ROCK.}\label{table:performKMGR}
\end{table}
\begin{table}[h!]
\begin{center}
\begin{footnotesize}
$$\begin{array}{c|c|ccccccccc}
&Train \ length & 2 & 5 & 10 & 15 & 25 & 50 & 75 & 300 & 600 \\\hline
&1    & 90 & 84 & 83 & 92 & 92 & 86 & 93 & 77 & 90\\
&2   & 84 & 86 & 84 & 84 & 82 & 87 & 92 & 79 & 71\\
&3  & 90  & 85 & 84 & 85 & 83 & 86 & 87 & 93 & 82\\
features&4  & 74 & 81 & 83 & 85 & 83 & 83 & 87 & 84 & 79\\
&5  & 90 & 90 & 85 & 89 & 83 & 85 & 86 & 71 & 85\\
&6  & 4 & 84 & 82 & 85 & 84 & 84 & 84 & 90 & 90\\
&7  & 90 & 90 & 89 & 88 & 90 & 84 & 84 & 85 & 90\\\hline
\end{array}$$
\end{footnotesize}
\end{center}
\caption{The validation performance (E1) of the ROCK model for each choice of trajectory length in the training set and test feature number for the double pendulum dataset. On the top left is the validation performance for ROCK with hyper-parameters consistent with the original MOCK implementation (1 feature and 2 sample training trajectories). Here the original MOCK will perform competitively. ROCK will often provide very small but performant models. In this case, one of the best performing models (300 sample training trajectories, with 3 test features) learns a model for the double pendulum with $\frac{1200}{300}\cdot 3 \cdot 2 = 24$ parameters, while the model for 600 sample training trajectories, with 1 test function has just 8 parameters but performs equal to or better than all but 5 of the models including MOCK, by contrast, the original MOCK model for this dataset has 2400 parameters.}\label{table:performdbpl}
\end{table}
\begin{table}[H]
\begin{center}
\begin{footnotesize}
$$\begin{array}{c|c|ccccccccc}
&Train \ length & 2 & 5 & 10 & 15 & 25 & 50 & 75 & 300 & 600 \\\hline
&1    & 39 & 37 & 37 & 33 & 32 & 39 & 30 & 41 & 7\\
&2   & 34 & 36 & 36 & 36 & 34 & 35 & 33 & 34 & 30\\
&3  & 18  & 42 & 40 & 36 & 41 & 32 & 30 & 36 & 29\\
features&4  & 35 & 41 & 36 & 50 & 34 & 35 & 37 & 35 & 34\\
&5  & 23 & 40 & 39 & 33 & 38 & 41 & 35 & 32 & 31\\
&6  & 35 & 30 & 36 & 37 & 46 & 38 & 41 & 31 & 35\\
&7  & 8 & 16 & 28 & 37 & 37 & 41 & 34 & 30 & 30\\\hline
\end{array}$$
\end{footnotesize}
\end{center}
\caption{The validation performance (E1) of the ROCK model for each choice of trajectory length in the training set and test feature number for the Lorenz  dataset. On the top left is the validation performance for ROCK with hyper-parameters consistent with the original MOCK implementation (1 feature and 2 sample training trajectories). Although the original MOCK score is good (39\%) there is another choice of parameters that performs much better (50\%).}\label{table:performLR}
\end{table}
\begin{table}[H]
\begin{center}
\begin{footnotesize}
$$\begin{array}{c|c|ccccccccc}
&Train \ length & 2 & 5 & 10 & 15 & 25 & 50 & 75 & 300 & 600 \\\hline
&1    & 98 & 98 & 98 & 98 & 99 & 96 & 93 & -5 & -75\\
&2   & 92 & 97 & 97 & 97 & 99 & 98 & 99 & 4 & -43\\
&3  & -2  & 2 & 88 & 87 & 98 & 93 & 98 & 82 & 34\\
features&4  & 16 & -54 & 61 & 85 & 98 & 89 & 99 & 96 & 23\\
&5  & -2 & -2 & -4 & 84 & 97 & 85 & 99 & 94 & 88\\
&6  & 9 & 4 & -52 & 26 & 82 & 89 & 100 & 95 & 94\\
&7  & 1 & -5 & 0 & -41 & 67 & 84 & 99 & 91 & 80\\
\end{array}$$
\end{footnotesize}
\end{center}
\caption{The validation performance (E1) of the ROCK model for each choice of trajectory length in the training set and test feature number for the Rossler  dataset. On the top left is the validation performance for ROCK with hyper-parameters consistent with the original MOCK implementation (1 feature and 2 sample training trajectories). Of note is the poor performance in the case where the training trajectories are short but the number of features in the test function space is large (this is an overdetermined regime and the linear system is ill conditioned). In addition performance degrades when the number of features is small and the trajectories are very long. This is the under determined regime because our final learned model has  a very small number of parameters. The model is too small to fit the data well in this regime.  }\label{table:performRS}
\end{table}

%\section[Proof of Thm]{Proof of \cref{thm:bigthm}}
%\label{sec:proof}
%\lipsum[106-112]

%\section{Additional experimental results}
%\Cref{tab:foo} shows additional
%supporting evidence. 

%\begin{table}[htbp]
%{\footnotesize
%  \caption{Example table}  \label{tab:foo}
%\begin{center}
%  \begin{tabular}{|c|c|c|} \hline
%   Species & \bf Mean & \bf Std.~Dev. \\ \hline
%    1 & 3.4 & 1.2 \\
%    2 & 5.4 & 0.6 \\ \hline
%  \end{tabular}
%\end{center}
%}
%\end{table}

\newpage

%%%%%%%%%%%%%%%%%%%%%%%%%%%%%%%%%%%%%%%%%%%%%%%%%%%%%%%%%%%%

\newpage
\section*{NeurIPS paper checklist}

\begin{enumerate}

\item {\bf Claims}
    \item[] Question: Do the main claims made in the abstract and introduction accurately reflect the paper's contributions and scope?
    \item[] Answer: \answerYes{} % Replace by \answerYes{}, \answerNo{}, or \answerNA{}.
    \item[] Justification: We begin by demonstrating a representer theorem \ref{thm:BigThm_sec} with many applications to ODEs and PDEs. In addition we apply this theorem for ODEs and PDEs creating the ROCK method which clearly demonstrates superior performance on many different datasets. This model also often has much fewer parameters than the comparator model MOCK. 

\item {\bf Limitations}
    \item[] Question: Does the paper discuss the limitations of the work performed by the authors?
    \item[] Answer: \answerYes{} % Replace by \answerYes{}, \answerNo{}, or \answerNA{}.
    \item[] Justification:  Our methods do not always outperform other methods. For example, our E2 score for KS for PDEs is rather poor for which we provide a brief discussion. In addition, ROCK is unable to outperform LODE for two of the datasets which is also mentioned.

\item {\bf Theory assumptions and proofs}
    \item[] Question: For each theoretical result, does the paper provide the full set of assumptions and a complete (and correct) proof?
    \item[] Answer: \answerYes{} % Replace by \answerYes{}, \answerNo{}, or \answerNA{}.
    \item[] Justification: Proofs of all theoretical results can be found in the appendices.

    \item {\bf Experimental result reproducibility}
    \item[] Question: Does the paper fully disclose all the information needed to reproduce the main experimental results of the paper to the extent that it affects the main claims and/or conclusions of the paper (regardless of whether the code and data are provided or not)?
    \item[] Answer: \answerYes{} % Replace by \answerYes{}, \answerNo{}, or \answerNA{}.
    \item[] Justification: We provide the code for running ROCK in the appendix, and additional code will be provided in later attachments. Our experimental sections also detail our validation procedures. Finally, we chose to benchmark with publicly available Common Task Framework data specifically to allow for comparison with other teams.

\item {\bf Open access to data and code}
    \item[] Question: Does the paper provide open access to the data and code, with sufficient instructions to faithfully reproduce the main experimental results, as described in supplemental material?
    \item[] Answer: \answerYes{} % Replace by \answerYes{}, \answerNo{}, or \answerNA{}.
    \item[] Justification: The CTF  datasets are publicly available upon registration to the Synapse website: \url{https://www.synapse.org/Synapse:syn52052735/wiki/622928}. We provide code for ROCK in the appendix and will provide code for the validation procedure when submitting additional material. 

\item {\bf Experimental setting/details}
    \item[] Question: Does the paper specify all the training and test details (e.g., data splits, hyperparameters, how they were chosen, type of optimizer, etc.) necessary to understand the results?
    \item[] Answer: \answerYes{} % Replace by \answerYes{}, \answerNo{}, or \answerNA{}.
    \item[] Justification: See Section \ref{sec:Experiments}. 

\item {\bf Experiment statistical significance}
    \item[] Question: Does the paper report error bars suitably and correctly defined or other appropriate information about the statistical significance of the experiments?
    \item[] Answer: \answerNo{} % Replace by \answerYes{}, \answerNo{}, or \answerNA{}.
    \item[] Justification: We did not have access to the test set of the CTF. Thus, we could not compute the statistical significance of the difference between methods. 

\item {\bf Experiments compute resources}
    \item[] Question: For each experiment, does the paper provide sufficient information on the computer resources (type of compute workers, memory, time of execution) needed to reproduce the experiments?
    \item[] Answer: \answerYes{} % Replace by \answerYes{}, \answerNo{}, or \answerNA{}.
    \item[] Justification: We provide information on computer resources in the captions of the tables reporting the experimental results.    
    
\item {\bf Code of ethics}
    \item[] Question: Does the research conducted in the paper conform, in every respect, with the NeurIPS Code of Ethics \url{https://neurips.cc/public/EthicsGuidelines}?
    \item[] Answer: \answerYes{} % Replace by \answerYes{}, \answerNo{}, or \answerNA{}.
    \item[] Justification: We have read the NeurIPS Code of Ethics and believe we are in full compliance with it.

\item {\bf Broader impacts}
    \item[] Question: Does the paper discuss both potential positive societal impacts and negative societal impacts of the work performed?
    \item[] Answer: \answerNA{} % Replace by \answerYes{}, \answerNo{}, or \answerNA{}.
    \item[] Justification: We have presented an algorithm for learning ODEs and PDEs from data.  We do not see any direct negative societal impacts of this work.
    
\item {\bf Safeguards}
    \item[] Question: Does the paper describe safeguards that have been put in place for responsible release of data or models that have a high risk for misuse (e.g., pretrained language models, image generators, or scraped datasets)?
    \item[] Answer: \answerNA{} % Replace by \answerYes{}, \answerNo{}, or \answerNA{}.
    \item[] Justification: In our opinion, the algorithms presented in this paper do not have a high risk for misuse.

\item {\bf Licenses for existing assets}
    \item[] Question: Are the creators or original owners of assets (e.g., code, data, models), used in the paper, properly credited and are the license and terms of use explicitly mentioned and properly respected?
    \item[] Answer: \answerYes{} % Replace by \answerYes{}, \answerNo{}, or \answerNA{}.
    \item[] Justification: We reference all the creators in the reference section. 

\item {\bf New assets}
    \item[] Question: Are new assets introduced in the paper well documented and is the documentation provided alongside the assets?
    \item[] Answer: \answerNA{} % Replace by \answerYes{}, \answerNo{}, or \answerNA{}.
    \item[] Justification: We do not release new assets.

\item {\bf Crowdsourcing and research with human subjects}
    \item[] Question: For crowdsourcing experiments and research with human subjects, does the paper include the full text of instructions given to participants and screenshots, if applicable, as well as details about compensation (if any)? 
    \item[] Answer: \answerNA{} % Replace by \answerYes{}, \answerNo{}, or \answerNA{}.
    \item[] Justification: We do not perform crowd sourcing and do not perform research on human subjects.

\item {\bf Institutional review board (IRB) approvals or equivalent for research with human subjects}
    \item[] Question: Does the paper describe potential risks incurred by study participants, whether such risks were disclosed to the subjects, and whether Institutional Review Board (IRB) approvals (or an equivalent approval/review based on the requirements of your country or institution) were obtained?
    \item[] Answer: \answerNA{} % Replace by \answerYes{}, \answerNo{}, or \answerNA{}.
    \item[] Justification: We do not perform crowd sourcing or research on human subjects.

\item {\bf Declaration of LLM usage}
    \item[] Question: Does the paper describe the usage of LLMs if it is an important, original, or non-standard component of the core methods in this research? Note that if the LLM is used only for writing, editing, or formatting purposes and does not impact the core methodology, scientific rigorousness, or originality of the research, declaration is not required.
    %this research? 
    \item[] Answer: \answerNA{} % Replace by \answerYes{}, \answerNo{}, or \answerNA{}.
    \item[] Justification: We do not use LLMs in this paper.

\end{enumerate}
\end{document}